\documentclass{article}

\usepackage[final, main]{neurips_2026}

\usepackage{todonotes}
\usepackage{amsthm}
\usepackage{xcolor}         
\definecolor{darkgreen}{rgb}{0.0, 0.5, 0.0} 
\definecolor{verylightgray}{rgb}{0.97, 0.97, 0.97}
\definecolor{darkorange}{rgb}{0.8, 0.4, 0.0}
\usepackage{wrapfig}
\usepackage{tcolorbox}
\usepackage{tabularx}
\newtcbox{\truthbox}[1][green!20!white]{on line, boxrule=0.5pt, colframe=green!50!black, colback=#1, sharp corners, boxsep=1pt, left=1pt, right=1pt, top=1pt, bottom=1pt}
\definecolor{verylightgray}{gray}{0.95}
\usepackage{tcolorbox}
\tcbuselibrary{skins}
\newtcbox{\wrongbox}{on line,
  colback=red!8, colframe=red!60,
  boxsep=0pt, left=2pt, right=2pt, top=1pt, bottom=1pt,
  boxrule=0.6pt, arc=2pt}
\newcommand{\cmark}{\textcolor{darkgreen}{\scalebox{1}[1.0]{\ding{51}}}}
\newcommand{\xmark}{\textcolor{red}{\ding{55}}}
\newcommand{\res}[2]{\textbf{#1} {\scriptsize(#2)}}
\definecolor{verylightgray}{gray}{0.95}
\newcommand{\best}[2]{\textbf{#1}{\color{gray}\scriptsize$\pm$#2}}

\usepackage[utf8]{inputenc} 
\usepackage[T1]{fontenc}    
\usepackage{hyperref}       
\usepackage{url}            
\usepackage{booktabs}       
\usepackage{amsfonts}       
\usepackage{nicefrac}       
\usepackage{microtype}      
\usepackage{amsmath}
\usepackage{enumitem}
\usepackage{pifont}
\usepackage{graphicx}
\usepackage{amsthm}

\usepackage{algorithm}
\usepackage{algpseudocode}
\usepackage[table]{xcolor} 
\usepackage{wrapfig}

\definecolor{cellSC}{HTML}{1F8A4F}\definecolor{cellAN}{HTML}{4C93D9}
\definecolor{cellSR}{HTML}{D55E00}\definecolor{cellCF}{HTML}{7A5195}
\newcommand{\cellbox}[1]{\textcolor{#1}{\rule{1.4ex}{1.4ex}}}
\renewcommand{\res}[2]{#1{\color{gray}\scriptsize$\pm$#2}}

\usepackage[utf8]{inputenc} 
\usepackage[T1]{fontenc}    
\usepackage{hyperref}       
\usepackage{url}            
\usepackage{booktabs}       
\usepackage{amsfonts}       
\usepackage{nicefrac}       
\usepackage{microtype}      
\usepackage{xcolor} 
\usepackage{booktabs}
\usepackage[table]{xcolor}
\usepackage{amsmath} 
\usepackage{enumitem}
\usepackage{booktabs}      
\usepackage{enumitem}      
\usepackage{amssymb}       
\usepackage{graphicx}      
\usepackage{textcomp}      
\usepackage{pifont}
\usepackage{xcolor}
\usepackage{pifont}
\usepackage{booktabs}
\usepackage{colortbl}   

\title{Detecting Outdated Knowledge\\ in Large Language Models}

\author{%
  \textbf{Rania Elbadry\textsuperscript{1} \quad
  Ahmed Heakl\textsuperscript{1} \quad
  Fan Zhang\textsuperscript{1} \quad
  Dani Bouch\textsuperscript{1}} \\
  \textbf{Yuxia Wang\textsuperscript{2} \quad
  Preslav Nakov\textsuperscript{1} \quad
  Zhuohan Xie\textsuperscript{1}} \\[0.5em]
  \textsuperscript{1}Mohamed bin Zayed University of Artificial Intelligence (MBZUAI) \\
  \textsuperscript{2}INSAIT, Sofia University ``St. Kliment Ohridski''\\
  \texttt{\{rania.elbadry, zhuohan.xie\}@mbzuai.ac.ae}
}

\usepackage{float}
\begin{document}

\maketitle
\begin{abstract}
Real-world facts, such as who leads a government, can change after a language model's training cutoff. As a result, a model may confidently repeat an outdated answer, making confidence-based checks insufficient for deciding when external verification is needed. Here we aim to study this. In particular, we investigate whether hidden states contain a signal of temporal knowledge drift that these checks fail to capture.
Using dated Wikidata timelines, we label 3,511 queries according to whether their answers have changed after each model's cutoff. We train a linear probe to predict drift and evaluate it on held-out facts. Across six instruction-tuned LLMs from four families, the probe detects drift with 0.80–0.93 AUROC, while prior methods using correctness, uncertainty, and consistency remain near chance (0.49–0.56). Moreover, the drift-probe direction is almost perpendicular to those of probes trained to detect correctness or uncertainty. In the models we test, the signal that predicts the drift is therefore encoded separately from whether the model is correct or unsure, which is why prior methods have missed it. Our results suggest that the models we tested encode whether their knowledge is likely out-of-date along an axis that confidence-based methods do not read. Our code and data are available at \url{https://github.com/mbzuai-nlp/geometry-of-staleness}.
\end{abstract}

\section{Introduction}\label{sec:intro}
\begin{wrapfigure}{r}{0.5\linewidth}
    \vspace{-1.5em}
    \centering
    \includegraphics[width=\linewidth]{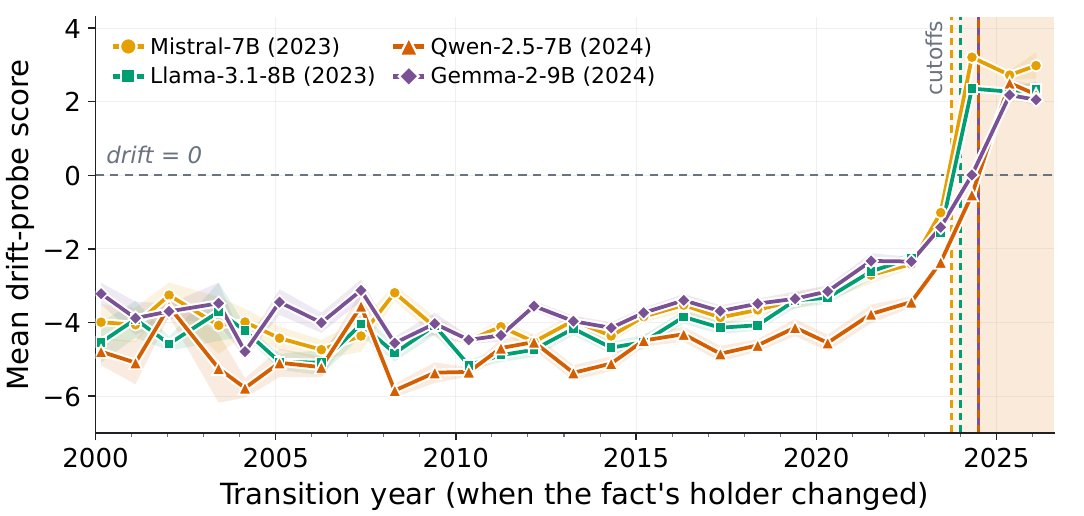}
    \vspace{-1.5em}
   \caption{Mean drift-probe score by year of recorded answer transition for four models. Vertical dashed lines mark the cutoff dates used in the analysis, while the horizontal dashed line marks the score of zero. Mean scores increase for transitions near and after the cutoffs.}
    \label{fig:teaser}
\end{wrapfigure}

Large language models (LLMs) can produce outdated answers as fluently and confidently as correct ones~\citep{mousavi-etal-2024-dyknow,vu2024freshllms,zhang-choi-2023-mitigating}. For example, a model trained on data from 2022 and asked who leads a company in 2025 may repeat the former CEO's name or invent an answer. These errors have different origins: the former answer reflects previously valid knowledge that has become outdated, while the latter is a fabrication. Existing detection methods~\citep{huang2025survey} generally flag factual errors without distinguishing between such cases. They rely on output-level statistics, such as entropy, top-$k$ token probabilities, and semantic clustering~\citep{farquhar2024detecting}, or on hidden-state probes trained to detect truthfulness or uncertainty~\citep{azaria2023internal,burns2022discovering,kossen2024semantic}. These signals do not directly target whether a fact has changed since the model's training cutoff.


\begin{figure}[t]
  \centering
  \includegraphics[width=\linewidth]{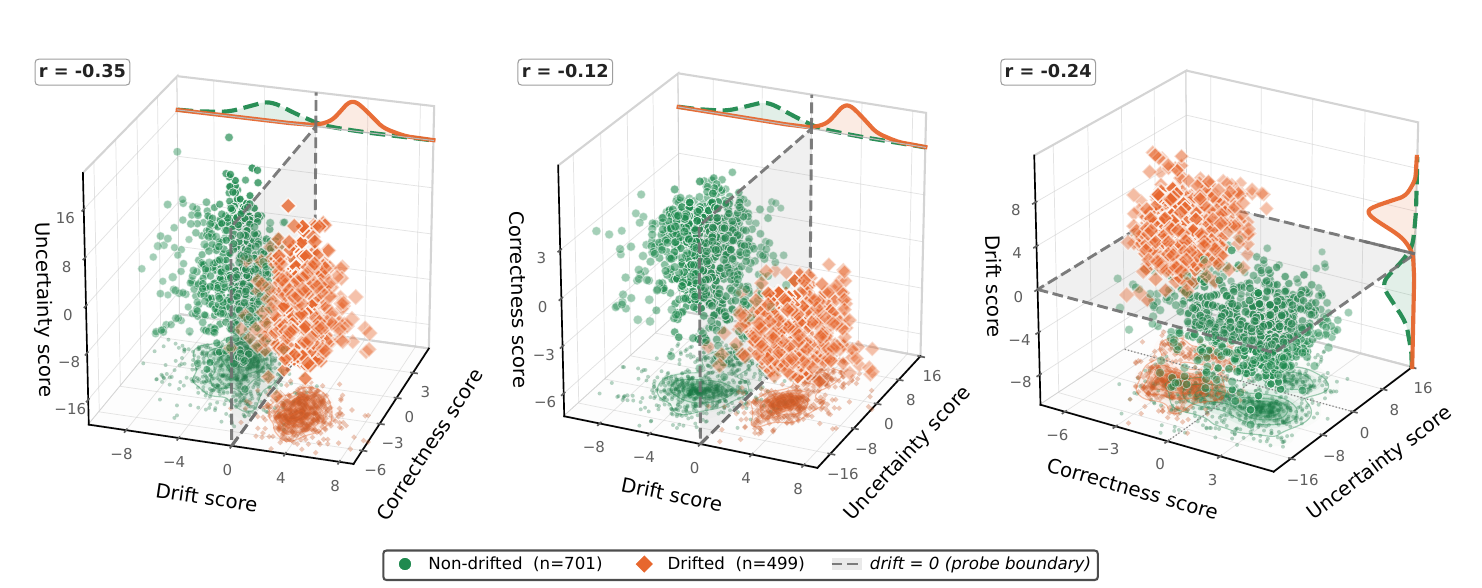}
  \vspace{-1.5em}
  \caption{Drift, correctness, and uncertainty probe scores for Mistral-7B at layer 6. The drift score distinguishes drifted from non-drifted facts more clearly (AUROC 0.99) than the correctness (0.71) or uncertainty (0.54) scores. Correctness and uncertainty have a weak negative correlation ($r$ = -0.24).}

  \label{fig:orthogonality}
  \vspace{-1.5em}
\end{figure}

These signals do not reliably distinguish outdated knowledge from other factual errors, especially when the model answers confidently. We therefore ask whether a model's hidden states contain a distinct signal of facts that changed after its training cutoff. We train a linear classifier, which we call the \emph{drift probe}, to predict temporal drift from these hidden states. Positive scores indicate predicted drift, while negative scores indicate no such drift. The mean score remains negative for fact transitions well before each model's cutoff and rises as transitions approach or pass it (Figure~\ref{fig:teaser}). Separate probes for answer correctness and uncertainty distinguish drifted from non-drifted facts less clearly (Figure~\ref{fig:orthogonality}).

Using dated Wikidata histories, we evaluate drift detection on held-out facts. Across six instruction-tuned models, methods based on output-level statistics (entropy, top-$k$ probability, and semantic clustering)~\citep{farquhar2024detecting} or hidden-state probes for truthfulness or uncertainty~\citep{azaria2023internal,burns2022discovering,kossen2024semantic} remain near chance, with $0.49$--$0.56$ AUROC, while our drift probe reaches $0.80$--$0.93$. Our contributions are as follows:
\begin{itemize}
\item \textbf{Dataset:} We construct a temporal question--answering dataset of 3,511 queries over 1,342 facts with dated answer histories. These histories provide model-specific drift labels and distinguish outdated answers from outputs that match no recorded answer.
\item \textbf{Method:} We develop a linear drift probe that achieves $0.80$--$0.93$ AUROC on six models. Its direction is almost perpendicular to those of correctness and uncertainty probes, suggesting that temporal drift is represented largely separately from these properties.
\item \textbf{Application:} We use the drift signal to prioritize external verification. In a simulation with a 20\% verification budget, the probe identifies 63\% of stale facts, compared with 49\% for the volatility-hazard baseline and 45\% for first-token entropy.
\end{itemize}

\section{Related Work}
\label{sec:related}

\paragraph{Temporal cutoffs and hallucination detection.}
Prior work has surveyed the causes, evaluation, and mitigation of
factual errors in LLMs~\citep{wang-etal-2024-factuality}. Benchmarks for time-stamped question--answering, such as TimeQA~\citep{chen2021dataset}, TempLAMA~\citep{dhingra-etal-2022-time}, StreamingQA~\citep{liska2022streamingqa}, and FreshLLMs~\citep{vu2024freshllms}, valuate temporal knowledge through model outputs, whereas we examine hidden states. Knowledge editing methods~\citep{meng2022locating,meng2022mass,mitchell2022memory} update outdated facts in the model weights, but require those facts to be identified first.. Hallucination detectors identify potentially incorrect answers using semantic entropy (SE)~\citep{kuhn2023semantic} or hidden-state probes, including semantic entropy probes
(SEP)~\citep{kossen2024semantic}, contrast-consistent search (CCS)~\citep{burns2022discovering}, and statement accuracy prediction based on language model activations (SAPLMA)~\citep{azaria2023internal}. Supervised uncertainty heads also detect hallucinations using attention maps and logits~\citep{shelmanov-etal-2025-head}.
These methods target uncertainty or factual correctness, whereas we predict whether a fact has changed since the model's training cutoff. Selective retrieval systems use model confidence, entity popularity, or model-generated retrieval requests to decide when to consult external sources~\citep{ICLR2024_25f7be96,jiang2023active,mallen2023not,xie-etal-2025-fire}. In contrast, we use drift scores to guide this decision.

\paragraph{Mechanistic interpretability and linear directions.}
Prior work has used linear probes to recover attributes from hidden states~\citep{alain2016understanding,tenney2019bert} and has shown that LLMs can represent concepts such as truth as linear directions~\citep{marks2023geometry,park2023linear}. Our drift probe follows this approach. Concept-erasure methods, including iterative null-space projection (INLP)~\citep{ravfogel2020null} and least-squares concept erasure (LEACE)~\citep{belrose2023leace}, can remove a linearly recoverable information about a chosen concept, such as gender, from the hidden states. We use them to remove inear information about correctness and uncertainty and measure how this affects drift detection. To study how models compute answers, prior work has used activation patching~\citep{heimersheim2024use,meng2022locating,wang2022interpretability}, direct logit attribution (DLA) via the logit lens~\citep{belrose2023eliciting,geva2022transformer,janiak2024adversarial}, and difference-of-means steering~\citep{subramani2022extracting,turner2023steering}, primarily to investigate factual recall, truthfulness, or sentiment. We apply these tools to \emph{temporal drift}, i.e., to changes in facts after an LLM's training cutoff. Makelov et al.~\cite{ICLR2024_70b8505a} have shown that changing an answer through steering does not prove that the model normally uses that direction. Similarly, our results point to intervention effects, not the model's use of the steering direction, which is distinct from the drift-probe direction.

\section{Problem Formulation and Dataset}
\label{sec:formulation}

\paragraph{Facts and temporal drift.}
\label{sec:drift_def}
We represent a fact $\phi=(e,r)$ as an entity $e$ paired with a relation $r$, such as \emph{United States} and \emph{head of government}. Its recorded answer history consists of $m$ non-overlapping intervals, $\bigl\{(h_i,t_{\text{start}}^{(i)},t_{\text{end}}^{(i)})\bigr\}_{i=1}^{m}$, where the \emph{holder} $h_i$ is the person or organization satisfying relation $r$ for entity $e$ during $[t_{\text{start}}^{(i)},t_{\text{end}}^{(i)})$. We denote the recorded answer at time $t$ by $a^*(\phi,t)$. A query $q=(\phi,t_q)$ asks for this answer at time $t_q$. We exclude queries with no recorded holder at $t_q$.
Let $\tau_\theta$ denote the cutoff date used for model $f_\theta$ in our analysis (Table~\ref{tab:models}), and let $t_{\text{now}}$ denote the fixed date of our Wikidata snapshot. We define \emph{temporal drift} by comparing the recorded answers at these two dates: $\textsc{Drifted}(\phi,\theta)\triangleq a^*(\phi,\tau_\theta)\neq a^*(\phi,t_{\text{now}})$. For example, a fact is drifted if its holder is $A$ at the cutoff and $B$ at the snapshot date, where $A \neq B$. This label applies regardless of whether the model learned the answer recorded at its cutoff. We evaluate answer correctness separately against $a^*(\phi,t_q)$. The model's output may match the holder at $t_q$, another recorded holder, or no holder in the history. Combining this match with the drift label yields the five answer categories in Table~\ref{tab:taxonomy}.

\begin{wraptable}{r}{0.52\linewidth}
\centering
\vspace{-1.8em}

\caption{Temporal knowledge states with examples from Llama-2-7B-Chat ($\tau_\theta\!=\!$~Sep~2022, snapshot Apr~2026). We define each cell by matching $f_\theta(q)$ against the holder timeline.  \emph{Correct}~indicates whether the output is factually correct at query time.}
\vspace{0.5em}
\label{tab:taxonomy}
\small
\setlength{\tabcolsep}{3pt}
\renewcommand{\arraystretch}{1.4}
\resizebox{\linewidth}{!}{%
\begin{tabular}{>{\centering\arraybackslash}m{2.8cm}
                >{\centering\arraybackslash}m{0.8cm}
                >{\centering\arraybackslash}m{1.2cm}
                m{4.5cm}}
\toprule
\textbf{Cell} & \textbf{Drift} & \textbf{Correct} & \textbf{Example} \\
\midrule
\rowcolor{verylightgray}
\textsc{Stable-Correct}
  & \xmark & \cmark
  & Who chairs Bank of England in 2014?
    \newline Says: Carney~~\truthbox{Truth: Carney} \\[4pt]
\rowcolor{white}
\textsc{Anachronism}
  & \xmark & \xmark
  & Who owns Dynabook in 2020?
    \newline Says: Toshiba~~\wrongbox{Truth: Sharp Corp.} \\[4pt]
\rowcolor{verylightgray}
\textsc{Stale-Recall}
  & \cmark & \xmark
  & Who coaches PSG in 2026?
    \newline Says: Tuchel~~\wrongbox{Truth: Luis Enrique} \\[4pt]
\rowcolor{white}
\textsc{Obsolete-Current}
  & \cmark & \xmark
  & Who governs Pennsylvania in 2024?
    \newline Says: Wolf~~\wrongbox{Truth: Shapiro} \\[4pt]
\rowcolor{verylightgray}
\textsc{Confabulation}
  & \cmark & \xmark
  & Who coaches NK Lokomotiva in 2026?
    \newline Says: Bili\'{c}~~\wrongbox{Truth: Jelavi\'{c}} \\
\bottomrule
\end{tabular}%
}
\end{wraptable}

\paragraph{Temporal knowledge states.}
\label{sec:taxonomy}
We normalize each model's output and match it to recorded holders using alias resolution. We combine the answer's correctness at $t_q$, the fact's drift label, and the matched holder's tenure relative to $\tau_\theta$ to define five categories~\citep{cheng2024dated}. For non-drifted facts, \textsc{Stable-Correct} denotes a correct answer at $t_q$, while \textsc{Anachronism} denotes an incorrect answer matching a holder whose tenure ended before $\tau_\theta$. For drifted facts, \textsc{Stale-Recall} denotes an incorrect answer matching a holder whose tenure ended before $\tau_\theta$, whereas \textsc{Obsolete-Current} denotes an incorrect answer matching the holder in office at $\tau_\theta$. \textsc{Confabulation} refers to an output for a drifted fact that matches no holder in the recorded timeline. For example, suppose a team was coached by $A$ until 2020, $B$ from 2020 to 2023, and $C$ from 2023 onward. Let both the query year and the reference year be 2026. With a 2024 cutoff, naming $C$ is \textsc{Stable-Correct}, while naming $A$ or $B$ is \textsc{Anachronism}. With a 2022 cutoff, naming $A$ is \textsc{Stale-Recall}, naming $B$ is \textsc{Obsolete-Current}, and naming an unrecorded coach $D$ is \textsc{Confabulation}.

\textsc{Stale-Recall} and \textsc{Obsolete-Current} match previously valid answers, whereas \textsc{Confabulation} denotes an output with no match in the recorded timeline. An unmatched output may result from an incomplete timeline or an alias matching error, and thus the label alone does not establish that the model invented the answer. We test whether a linear probe can distinguish \textsc{Stale-Recall} from \textsc{Confabulation} in Section~\ref{sec:decodability}. Other outcomes, including correct answers on drifted facts, and the complete assignment procedure are described in Appendix~\ref{app:cell_assignment}.

\paragraph{Dataset.}
Prior temporal question-answering datasets~\citep{dhingra-etal-2022-time,liska2022streamingqa} provide answers to individual queries but lack the complete holder histories needed to assign our temporal knowledge states. We therefore construct a dataset of $3{,}511$ queries over $1{,}342$ facts, covering $1{,}334$ entities and $4$ relations. Each query pairs an entity and a relation with a query year between 2010 and 2026. We select four relations, \texttt{head\_of\_government}, \texttt{head\_coach}, \texttt{chair\_of}, and \texttt{owned\_by}, because their holders change frequently, Wikidata records start and end dates for them, and they cover governance, sports, corporate leadership, and ownership. For each fact, we retrieve its complete holder history from our Wikidata snapshot. Most queries concern recent years. We retain only years in which a single holder occupied the role throughout the year, ensuring that every query has a single correct answer.

\textbf{Probing setup.}
\label{sec:probes}We evaluate six instruction-tuned models from four architecture
families with training cutoffs from September~2022 to June~2024
(Table~\ref{tab:models}).  For each model we extract the
residual-stream activation
$\mathbf{h}^{(\ell)} \in \mathbb{R}^d$ at every layer $\ell$ at the generated answer's first content-token position (the second token position for Llama-2, which emits a leading whitespace token).  We train $L_1$-regularized linear probes using the
Iterative Shrinkage-Thresholding Algorithm (ISTA),
\begin{equation}
\label{eq:probe}
\hat{y}_d = \sigma\!\bigl(\mathbf{w}_d^\top
\mathbf{h}^{(\ell)} + b_d\bigr),
\quad
\mathcal{L} = \mathrm{BCE}(\hat{y}_d, y_d)
  + \lambda \|\mathbf{w}_d\|_1,
\end{equation}
\begin{wraptable}{r}{0.42\linewidth}
\centering
\vspace{-1em}

\caption{Drift probe results. With all years, the year hints at drift; post-cutoff years remove this.}
\vspace{0.5em}

\label{tab:drift_probes}
\scriptsize
\rowcolors{2}{verylightgray}{white}
\setlength{\tabcolsep}{2pt}
\begin{tabular}{lcccc}
\toprule
 & \multicolumn{2}{c}{\textbf{All years}}
 & \multicolumn{2}{c}{\textbf{Post-cutoff years}} \\
\cmidrule(lr){2-3} \cmidrule(lr){4-5}
\textbf{Model} & \textbf{Layer} & \textbf{AUROC} & \textbf{Layer} & \textbf{AUROC} \\
\midrule
Llama-2   & L30 & 0.975\,$\pm$\,0.010 & L28 & 0.914\,$\pm$\,0.013 \\
Mistral   & L6  & 0.973\,$\pm$\,0.011 & L6  & 0.925\,$\pm$\,0.012 \\
Llama-3.1 & L18 & 0.959\,$\pm$\,0.016 & L27 & 0.923\,$\pm$\,0.010 \\
Qwen-2.5  & L8  & 0.961\,$\pm$\,0.015 & L22 & 0.827\,$\pm$\,0.015 \\
Gemma-9B  & L10 & 0.961\,$\pm$\,0.014 & L28 & 0.801\,$\pm$\,0.013 \\
Gemma-2B  & L10 & 0.942\,$\pm$\,0.019 & L16 & 0.828\,$\pm$\,0.012 \\
\bottomrule
\end{tabular}
\vspace{-1.0em}
\end{wraptable}
where $\mathbf{w}_d$ and $b_d$ are the weights and bias of the probe, $\sigma$ is the sigmoid, $y_d$ is the label, BCE is binary cross-entropy, and $\lambda$ controls the strength of the $L_1$ penalty. 

We split the data by fact, so all queries about the same (entity, relation) pair remain in one split, with 80\% of facts used for training and 20\% for testing. If queries about the same fact appeared in both splits, the probe could memorize its label instead of learning to detect drift.
Because query year and drift are correlated, a probe could predict drift from the year alone. To reduce this shortcut, we train and evaluate only on query years after each model's cutoff year, a setting we call the \emph{controlled evaluation}. We jointly select the layer and $\lambda \in \{10^{-5}, 10^{-4}, 10^{-3}, 10^{-2}, 10^{-1}\}$ using 3-fold cross-validation on the training facts, again grouped by fact. Drifted and non-drifted queries are not equally common, so we balance the training classes by randomly oversampling the smaller class. Without this step, the probe could favor the majority class. After selecting the layer and $\lambda$, we train the final probe on all training facts and evaluate it once on the held-out test facts, which are never used for model selection.

\section{Linear Decodability of Temporal Drift}
\begin{wrapfigure}{r}{0.45\linewidth}
    \centering
    \vspace{-1.8em}
    \includegraphics[width=\linewidth,trim=0 15 15 15, clip]{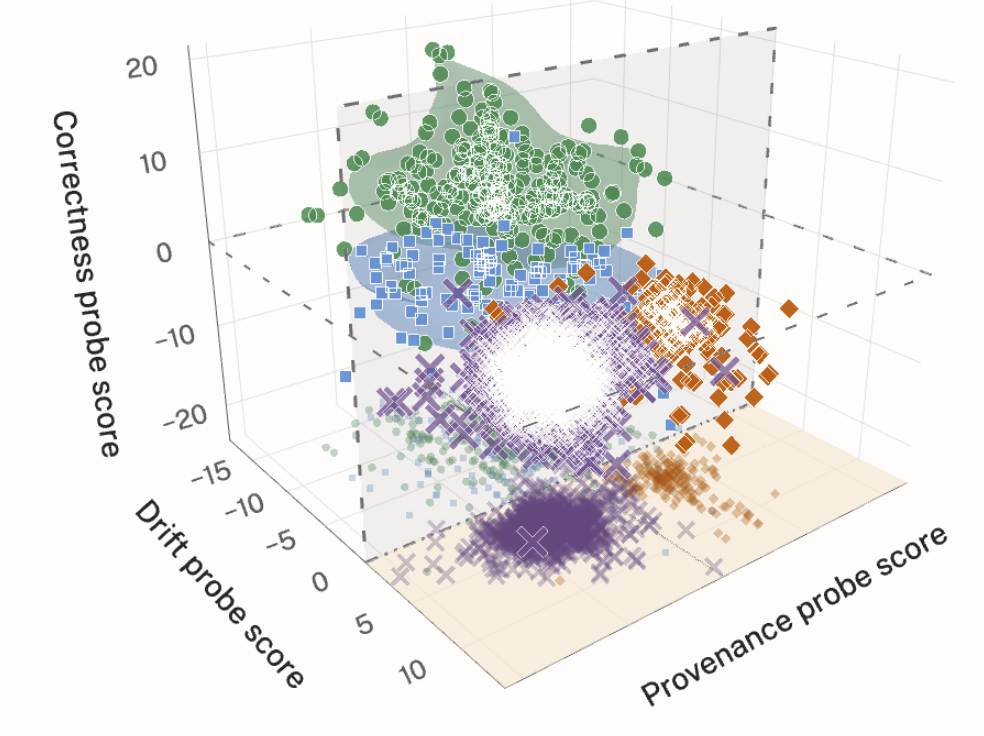}
    \caption{Drift and provenance axes are jointly
    decodable (Llama-2, controlled layer).  A probe
    separates \cellbox{cellSR}\,\textsc{Stale-Recall} from \cellbox{cellCF}\,\textsc{Confabulation}
    at AUROC\,=\,0.96 within the drifted class
    (Table~\ref{tab:provenance_probe}).
    \cellbox{cellSC}\,\textsc{Stable-Correct} and \cellbox{cellAN}\,\textsc{Anachronism} (non-drifted
    cells) cluster at negative drift scores.}
    \vspace{-1.9em}
    \label{fig:subcell_scatter}
\end{wrapfigure}

\label{sec:decodability}
Our drift probe achieves an AUROC of $0.80$--$0.93$ across six models (Table~\ref{tab:drift_probes}). This shows that model hidden states contain linearly decodable information about whether a fact has changed since the model's training cutoff. To test generalization to unseen relations, we extend the dataset from 4 to 18 relations. Training on 17 relations and testing on the remaining one yields a mean AUROC of $0.821$ (Appendix~\ref{app:relations}).

We report results on post-cutoff query years only, because query year can provide a shortcut for predicting drift. Drifted queries mostly concern years after the cutoff. When training includes all years, the probe may exploit the year in the prompt rather than the model's knowledge. Table~\ref{tab:drift_probes} illustrates this. With all years, AUROC is $0.94$--$0.98$. Restricting evaluation to post-cutoff years reduces AUROC by $0.09$ on average, consistent with a contribution from the year shortcut. The selected layer changes for five of six models, with the largest shift on Gemma-2-9B, from layer 10 to layer 28 (Appendix~\ref{app:layer_shifts}).

With post-cutoff query years, the signal is not confined to a single layer: the top five layers are within $0.007$ AUROC of one another on every model. We further test the drift probe on two groups of queries where the model names a previous holder, but only one group concerns drifted facts. In \textsc{Stale-Recall}, the fact has changed since the cutoff, whereas in \textsc{Anachronism}, it has not. If the probe responded only to old names, it should score both groups similarly. Instead, it scores \textsc{Stale-Recall} higher than \textsc{Anachronism} on every model (Appendix~\ref{app:anachronism_falsifier}), supporting the interpretation that it detects changes in the fact rather than merely the presence of an old name.

\paragraph{Outdated versus fabricated answers.}
\label{sec:provenance}
The previous test compared two groups that name past holders but differ in whether the fact has drifted. We now compare two groups whose facts have both drifted but whose incorrect answers have different origins. In \textsc{Stale-Recall}, the model names a recorded holder whose term ended before its cutoff. In \textsc{Confabulation}, its answer matches no holder in the timeline (Section~\ref{sec:taxonomy}). Both are factual errors, but one matches a previously valid answer and the other does not. To test whether hidden states distinguish them, we train a second, \emph{provenance probe}, on these two groups alone. It achieves $0.89$--$0.99$ AUROC across the six models (mean $0.96$, Table~\ref{tab:provenance_probe}).
Figure~\ref{fig:subcell_scatter} plots three probe scores for Llama-2, with one point per query. Along the drift axis, the non-drifted groups, \textsc{Stable-Correct} and \textsc{Anachronism}, have negative scores, while the drifted groups have positive scores. Along the provenance axis, \textsc{Stale-Recall} and \textsc{Confabulation} form distinct clusters. Along the correctness axis, \textsc{Stable-Correct}, the only correct-answer group shown, scores highest. Thus, the same hidden states contain information about both whether a fact has changed and whether an incorrect answer matches a previously valid holder.

\section{Geometric Independence of Temporal Drift}
\label{sec:orthogonality}
\begin{wraptable}{r}{0.25\linewidth}
\scriptsize
\centering
\vspace{-2.3em}
\caption{\textsc{Stale-Recall} vs.\ \textsc{Confabulation} probe. Selected-layer AUROC ($\pm$ 95\% CI half-width) on the controlled test set.}
\vspace{0.5em}
\label{tab:provenance_probe}
\rowcolors{2}{verylightgray}{white}
\begin{tabular}{lc}
\toprule
\textbf{Model} & \textbf{AUROC} \\
\midrule
Llama-2   & 0.965\,$\pm$\,0.027 \\
Mistral   & 0.990\,$\pm$\,0.015 \\
Llama-3.1 & 0.987\,$\pm$\,0.017 \\
Qwen-2.5  & 0.889\,$\pm$\,0.111 \\
Gemma-2-9B  & 0.939\,$\pm$\,0.070 \\
Gemma-2-2B  & 0.995\,$\pm$\,0.009 \\
\bottomrule
\end{tabular}
\vspace{-1.5em}
\end{wraptable}

Section~\ref{sec:decodability} leaves open whether drift is a distinct signal or merely a linear combination of correctness and uncertainty. In the latter case, the drift probe would provide no additional information.

\paragraph{Weight cosines.}
\label{sec:weight_cosines}
We first examine the angles between the probe weights. Let $\mathbf{w}_d$, $\mathbf{w}_c$, and $\mathbf{w}_u$ denote the weight vectors of the drift, correctness, and uncertainty probes. Alignment between $\mathbf{w}_d$ and either reference direction would yield a large absolute cosine. However, our $L_1$ penalty encourages sparse weights, and sparse vectors can have near-zero cosine simply because they use different dimensions. We therefore retrain all three probes with an $L_2$ penalty, which does not enforce sparsity, at each model's selected layer. All 12 absolute cosines between the drift probe and the two reference probes, two per model, are at or below 0.090, with a median of 0.025 (Table~\ref{tab:orthogonality_main}). Thus, the drift direction is nearly perpendicular to both reference directions in every model. The $L_2$ probes also achieve AUROC within 0.005 of the $L_1$ probes at the same layers, indicating that both predictive performance and near-orthogonality persist without the sparsity-inducing penalty.
\paragraph{Score correlations.}
\label{sec:score_corr}
Weight cosines alone do not establish whether probes produce independent scores. Even nearly orthogonal probes can yield correlated scores if hidden states vary more along some directions than others. We therefore measure correlations between probe scores on the test set. Drift and correctness scores are weakly correlated, with $r$ ranging from $-0.12$ to $-0.02$ (Table~\ref{tab:orthogonality_main}). The correlation between drift and uncertainty is also small, with $|r| \leq 0.20$. By comparison, correctness and uncertainty correlate more strongly with each other, with $r$ ranging from $-0.40$ to $-0.13$ (Appendix~\ref{app:cu_correlations}), consistent with models being less certain when their answers are wrong. Drift scores show little association with either reference score.

\paragraph{Null-space projection.}
\label{sec:nullspace}
Weak pairwise correlations do not rule out the possibility that drift is captured by a linear combination of correctness and uncertainty. To test this, we project both probe directions out of the hidden states and retrain the drift probe. More generally, for each target probe $t \in \{d, c, u\}$, let $\mathbf{W}_{-t}$ contain the unit-normalized $L_2$ weight vectors of the other two probes as columns. We remove their joint span from each hidden state $\mathbf{h}$ using the standard orthogonal projection~\citep{belrose2023leace,ravfogel2020null}:
\begin{equation}
\label{eq:nullspace}
\mathbf{h}_{\perp}^{(t)} = \mathbf{h} - \mathbf{W}_{-t}\bigl(\mathbf{W}_{-t}^{\top}\mathbf{W}_{-t}\bigr)^{-1}\mathbf{W}_{-t}^{\top}\,\mathbf{h}.
\end{equation}

\begin{wraptable}{r}{0.48\linewidth}
\centering
\vspace{-0.5em}
\caption{Overlap of the drift probe with the correctness and uncertainty probes. \textit{Cosine}: absolute cosine between dense $L_2$ probe weights at the selected layer. }
\label{tab:orthogonality_main}\label{tab:weight_score_indep}
\scriptsize
\rowcolors{2}{verylightgray}{white}
\setlength{\tabcolsep}{2.5pt}
\renewcommand{\arraystretch}{1.1}
\begin{tabular}{lcccc}
\toprule
 & \multicolumn{2}{c}{Drift--correctness}
 & \multicolumn{2}{c}{Drift--uncertainty} \\
\cmidrule(lr){2-3} \cmidrule(lr){4-5}
\textbf{Model} & Cosine & Correlation & Cosine & Correlation \\
\midrule
Llama-2    & 0.086 & $-0.02\pm0.08$ & 0.011 & $+0.11\pm0.08$ \\
Mistral    & 0.015 & $-0.07\pm0.08$ & 0.018 & $-0.09\pm0.08$ \\
Llama-3.1  & 0.090 & $-0.02\pm0.08$ & 0.050 & $-0.13\pm0.08$ \\
Qwen-2.5   & 0.023 & $-0.12\pm0.08$ & 0.008 & $+0.12\pm0.08$ \\
Gemma-2-9B & 0.050 & $-0.11\pm0.08$ & 0.026 & $+0.13\pm0.09$ \\
\rowcolor{white}
Gemma-2-2B & 0.024 & $-0.03\pm0.09$ & 0.035 & $+0.20\pm0.09$ \\
\bottomrule
\end{tabular}
\end{wraptable}

This projection removes the span of both directions, even when they are not mutually orthogonal. We then train a new $L_2$ target probe on $\mathbf{h}_{\perp}^{(t)}$. If target information were confined to that span, detection performance would fall to chance. Instead, across all 18 combinations of six models and three targets, AUROC changes by at most 0.008, with a median change of 0.0006. For drift, the change is at most 0.002 on every model (Table~\ref{tab:subspace_tests}). To contextualize these changes, we also remove two random directions. The 95th percentile of the resulting AUROC change ranges from 0.0002 to 0.0006 across models, and the drift change falls within the corresponding random-projection threshold on three of six models (Appendix~\ref{app:nullspace}). Thus, removing the correctness and uncertainty directions barely affects drift detection. Each of the other two targets is similarly robust to removing the remaining probe directions.


\textbf{Iterative null-space projection.}
\label{sec:inlp}
The preceding projection removes only one direction per signal. If correctness or uncertainty is encoded across multiple directions, a single projection may leave some information intact. Iterative null-space projection (INLP)~\citep{ravfogel2020null} addresses this possibility by repeatedly training a linear classifier for the signal to be removed and projecting its weight direction out of the hidden states:
\begin{equation}
\label{eq:inlp}
\mathbf{h}^{(i+1)} = \mathbf{h}^{(i)} - \mathbf{w}_i\bigl(\mathbf{w}_i^{\top}\mathbf{w}_i\bigr)^{-1}\mathbf{w}_i^{\top}\,\mathbf{h}^{(i)}, \quad i = 1, \ldots, k.
\end{equation}
After $k = 10$ steps, we retrain the target probe on $\mathbf{h}^{(k+1)}$ and measure the change in AUROC. For example, we retrain the drift probe after removing ten correctness directions. As a baseline, we repeat the procedure 100 times, removing 10 random directions in each run.

\begin{wraptable}{r}{0.52\linewidth}
\centering
\vspace{-1.5em}
\caption{Removing correctness and uncertainty barely changes drift detection. \textit{Null-space}: change in drift AUROC after projecting out the correctness and uncertainty directions (Eq.~\ref{eq:nullspace}). \textit{INLP}: change in drift AUROC after removing 10 directions with INLP. P95 is the 95th percentile of the change when removing 10 random directions. All changes are at most 0.006.}
\vspace{0.5em}
\label{tab:subspace_tests}
\scriptsize
\rowcolors{2}{verylightgray}{white}
\resizebox{\linewidth}{!}{
\begin{tabular}{lccccc}
\toprule
\textbf{Model} & \shortstack{\textbf{Null-}\\\textbf{space}}
  & \multicolumn{2}{c}{\textbf{Drift--correctness}}
  & \multicolumn{2}{c}{\textbf{Drift--uncertainty}} \\
\cmidrule(lr){3-4} \cmidrule(lr){5-6}
 & $|\Delta|$
 & $\Delta$ & P95
 & $\Delta$ & P95 \\
\midrule
Llama-2    & 0.001 & $+$0.006 & 0.001 & 0.000 & 0.001 \\
Mistral    & 0.000 & $-$0.001 & 0.001 & $-$0.002 & 0.001 \\
Llama-3.1  & 0.002 & $+$0.005 & 0.001 & 0.000 & 0.001 \\
Qwen-2.5   & 0.001 & $-$0.001 & 0.001 & 0.000 & 0.001 \\
Gemma-2-9B & 0.000 & $-$0.004 & 0.002 & $+$0.005 & 0.002 \\
\rowcolor{white}
Gemma-2-2B & 0.001 & $+$0.003 & 0.001 & 0.000 & 0.001 \\
\bottomrule
\end{tabular}
}
\vspace{-1.0em}
\end{wraptable}
Removing ten uncertainty directions changes drift AUROC by at most 0.005 across six models, indicating limited overlap. Removing ten correctness directions changes it by at most 0.006, below every drift AUROC's confidence-interval half-width in Table~\ref{tab:drift_probes}. Some changes exceed the 95th percentile of the random baseline, but remain below 0.01. The largest changes occur on Llama-2 and Llama-3.1, where the drift and correctness probes use nearby layers. Correctness and uncertainty show greater overlap: INLP changes AUROC by up to 0.020 on three of six models, consistent with lower certainty for incorrect answers. Drift detection remains largely unaffected.

\textbf{Mean-shift directions.}
\label{sec:dom_dissociation}
Earlier results could reflect probe training. We use difference-of-means (DoM) directions $\boldsymbol{\mu}$, computed without classifier training. Each direction is the difference between two groups' mean hidden states, such as drifted and non-drifted queries. These directions overlap more than trained probe directions. Correctness and uncertainty DoM directions have cosines from $-0.92$ to $-0.37$, whereas corresponding trained probes have absolute cosines at or below 0.15. Similarly, drift's DoM direction overlaps with both correctness and uncertainty ($|\cos| = 0.09$--$0.60$), while its trained probe direction is nearly perpendicular to both ($|\cos| \leq 0.08$).
This difference is consistent with overlap between the groups: most drifted queries receive incorrect answers, so mean differences can mix features associated with multiple signals~\citep{patel2026llm,cho2026confidence}. Regularized probes instead learn weights that discriminate the target labels~\citep{patel2026llm,miao2026closing}. Full results appear in Table~\ref{tab:dom_vs_probe} in Appendix~\ref{app:dom_dissociation}. Across the five tests, drift remains linearly decodable beyond the correctness and uncertainty directions, indicating that readouts restricted to those directions miss useful drift information. This conclusion is limited to linear readouts of the hidden states studied here. Nonlinear methods, other representations, and alternative uncertainty probes remain to be tested.

\begin{table}[t]
\centering
\caption{Drift detection AUROC on the controlled
(post-cutoff) split.  The drift-supervised probes are trained on drift labels, while the other methods target correctness, uncertainty, or consistency, so the table tests the prediction from
Section~\ref{sec:orthogonality} rather than ranking them.  Non-drift methods reach at most 0.56
mean AUROC.}
\label{tab:baselines}
\small
\setlength{\tabcolsep}{4pt}
\resizebox{\linewidth}{!}{
\begin{tabular}{llccccccr}
\toprule
& & \textbf{Llama-2} & \textbf{Mistral} & \textbf{Llama-3.1}
  & \textbf{Qwen-2.5} & \textbf{Gemma-9B} & \textbf{Gemma-2B} & \textbf{Mean} \\
\midrule
\rowcolor{white}
\multicolumn{9}{l}{\textit{Output-based}} \\
\rowcolor{verylightgray}
& Token entropy          & \res{0.52}{0.05} & \res{0.50}{0.07} & \res{0.57}{0.08} & \res{0.47}{0.12} & \res{0.55}{0.11} & \res{0.57}{0.12} & \res{0.53}{0.03} \\
\rowcolor{white}
& Token probability      & \res{0.51}{0.05} & \res{0.49}{0.07} & \res{0.55}{0.08} & \res{0.47}{0.12} & \res{0.54}{0.11} & \res{0.54}{0.12} & \res{0.52}{0.03} \\
\rowcolor{verylightgray}
& Sequence log-probability & \res{0.50}{0.05} & \res{0.55}{0.07} & \res{0.47}{0.07} & \res{0.47}{0.12} & \res{0.51}{0.11} & \res{0.55}{0.12} & \res{0.51}{0.03} \\
\rowcolor{white}
& Sequence entropy       & \res{0.51}{0.05} & \res{0.54}{0.07} & \res{0.47}{0.07} & \res{0.47}{0.12} & \res{0.50}{0.11} & \res{0.56}{0.12} & \res{0.51}{0.03} \\
\rowcolor{verylightgray}
& Top-$k$ margin         & \res{0.51}{0.05} & \res{0.49}{0.07} & \res{0.51}{0.07} & \res{0.46}{0.12} & \res{0.52}{0.11} & \res{0.44}{0.12} & \res{0.49}{0.03} \\
\rowcolor{white}
& Generation length      & \res{0.50}{0.05} & \res{0.52}{0.07} & \res{0.63}{0.08} & \res{0.47}{0.12} & \res{0.60}{0.11} & \res{0.50}{0.12} & \res{0.54}{0.03} \\
\rowcolor{verylightgray}
& Semantic entropy~\citep{farquhar2024detecting}
                         & \res{0.52}{0.05} & \res{0.53}{0.07} & \res{0.54}{0.08} & \res{0.53}{0.12} & \res{0.55}{0.11} & \res{0.55}{0.12} & \res{0.54}{0.03} \\
\rowcolor{white}
& Scalar ensemble        & \res{0.53}{0.05} & \res{0.49}{0.07} & \res{0.62}{0.08} & \res{0.47}{0.12} & \res{0.54}{0.11} & \res{0.50}{0.12} & \res{0.53}{0.03} \\
\midrule
\rowcolor{white}
\multicolumn{9}{l}{\textit{Representation-based}} \\
\rowcolor{verylightgray}
& SEP~\citep{kossen2024semantic}
                         & \res{0.53}{0.05} & \res{0.54}{0.07} & \res{0.55}{0.08} & \res{0.54}{0.12} & \res{0.56}{0.11} & \res{0.65}{0.12} & \res{0.56}{0.03} \\
\rowcolor{white}
& SAPLMA~\citep{azaria2023internal}
                         & \res{0.54}{0.05} & \res{0.50}{0.07} & \res{0.61}{0.08} & \res{0.59}{0.12} & \res{0.52}{0.11} & \res{0.53}{0.12} & \res{0.55}{0.03} \\
\rowcolor{verylightgray}
& CCS~\citep{burns2022discovering}
                         & \res{0.54}{0.05} & \res{0.51}{0.07} & \res{0.53}{0.08} & \res{0.54}{0.12} & \res{0.55}{0.11} & \res{0.53}{0.12} & \res{0.53}{0.03} \\
\midrule
\rowcolor{white}
\multicolumn{9}{l}{\textit{Drift-supervised}} \\
\rowcolor{verylightgray}
& MLP probe (drift)      & \res{0.89}{0.03} & \res{0.92}{0.04} & \res{0.92}{0.04} & \res{0.63}{0.12} & \res{0.75}{0.10} & \res{0.78}{0.11} & \res{0.82}{0.04} \\
\rowcolor{white}
& \textbf{Our probe} & \best{0.91}{0.01} & \best{0.93}{0.01} & \best{0.92}{0.01} & \best{0.83}{0.02} & \best{0.80}{0.01} & \best{0.83}{0.01} & \best{0.87}{0.01} \\
\bottomrule
\end{tabular}
}
\end{table}
\section{Consequences: Detection Failure and Model Specificity}

\label{sec:consequences}
\begin{wrapfigure}{r}{0.50\linewidth}
    \centering
    \vspace{-1em}
    \includegraphics[width=\linewidth]{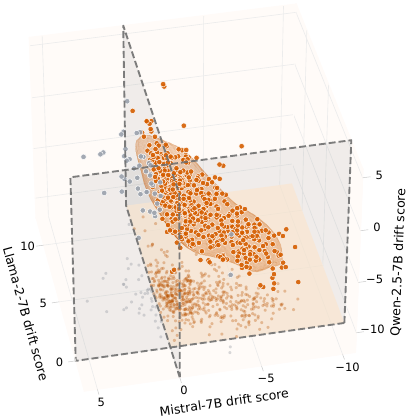}
\caption{Cross-cutoff drift probe scores on identical inputs for three models (Table~\ref{tab:hero}). Each point is one query, and the earlier-cutoff model scores higher on facts whose holder changed between the cutoffs, so the score depends on the specific model.}  
\label{fig:hero_scatter}
    \vspace{-1.5em}
\end{wrapfigure}

Section~\ref{sec:orthogonality} has shown limited overlap between drift, correctness, and uncertainty, suggesting that methods based on the latter two signals may struggle to detect drift. Table~\ref{tab:baselines} supports this on post-cutoff query years. Output-based methods, including token entropy and semantic entropy~\citep{farquhar2024detecting}, achieve mean AUROC of 0.49--0.54, while hidden-state probes achieve 0.53--0.56: CCS~\citep{burns2022discovering} reaches 0.53, SAPLMA~\citep{azaria2023internal} 0.55, and SEP~\citep{kossen2024semantic} 0.56. Our drift probe achieves 0.87. Most of this gap comes from \textsc{Stale-Recall}, where the model confidently recalls a previously correct answer. An entropy threshold at the 80th percentile misses 55\% of these answers, and 31\% have lower entropy than the median correct answer (Table~\ref{tab:asymmetric}).

\paragraph{No-internals baseline.}
\label{sec:hazard}
The strongest baseline uses no model internals. The volatility-hazard score uses a constant-rate hazard model from survival analysis~\citep{cox1972regression,kalbfleisch2002statistical} to predict drift based on the recorded training pre-cutoff change frequency: $s = 1 - \exp[-\rho\,(t_q - \tau_\theta)]$, where $\rho$ is the annual rate of holder changes before $\tau_\theta$. On the 92\% of queries whose facts changed before the cutoff, it achieves 0.709 AUROC, outperforming all correctness- and uncertainty-based methods in Table~\ref{tab:baselines}. Fact history thus outperforms these model-derived signals. This challenges our probe, which could learn how often facts change rather than detect information specific to the model's knowledge. The probe achieves 0.905 AUROC on the same queries (Appendix~\ref{app:hazard}), indicating predictive information beyond historical change rates. On the remaining 8\% with no pre-cutoff changes the hazard falls to chance (0.500), while the probe achieves 0.777. Moreover, performance is lower for facts rarely mentioned with dates in pretraining: AUROC falls from 0.945 for the quarter of facts with the most dated mentions in Dolma~\citep{soldaini2024dolma} to 0.782 for the quarter with the fewest (Appendix~\ref{app:coverage}), suggesting that drift is more detectable for facts with greater temporal evidence in pretraining.

\paragraph{Model specificity.}
\label{sec:hero}
A remaining concern is that the probe detects input properties, such as entity familiarity, relation type, or the gap between the query year and the cutoff, rather than information about the model's knowledge. We test this possibility with a cross-cutoff experiment using identical inputs. For facts whose holder changed \emph{between} two models' cutoffs, the fact has drifted for the earlier-cutoff model $A$ but not the later-cutoff model $B$. We give both models identical prompts, apply their respective drift probes, and measure $P(A > B)$, the probability that model $A$ receives a higher score than model $B$. Across all pairs, $P(A > B)$ ranges from 0.975 to 0.998 (see Table~\ref{tab:hero} in Appendix~\ref{app:pair_independence}), covering four architecture families and cutoff gaps as short as 3 months (see Figure~\ref{fig:hero_scatter}). The reverse pattern occurs on only 0.0--0.2\% of queries. Because the prompts are identical, fixed input properties alone cannot explain the score differences, supporting a model-specific interpretation. This test does not establish which aspects of training produce the signal.
A complementary test holds the model fixed and varies the prompt. A probe trained on the original prompt shows an average AUROC change of only 0.034 across four unseen prompt formats and retains 91\% of its margin above chance when temporal words are removed (Appendix~\ref{app:prompt_robustness}). These results indicate that the signal is not tied to a single prompt format. Section~\ref{sec:mechanism} examines \emph{how} the model computes and encodes it.

\begin{wraptable}{r}{0.5\linewidth}
\centering
 \vspace{-1.7em}
\caption{Cross-cutoff experiment on identical inputs. The drift probe gives a higher score to the model whose cutoff comes before the holder change. The Mann--Whitney test is one-sided.}
\label{tab:hero}
\vspace{0.5em}
\scriptsize
\setlength{\tabcolsep}{4pt}
\rowcolors{2}{verylightgray}{white}
\renewcommand{\arraystretch}{1.5}
\resizebox{\linewidth}{!}{
\begin{tabular}{llcccr}
\toprule
\textbf{Model A} & \textbf{Model B} & \shortstack{\textbf{Matched}\\\textbf{queries}}
  & $\boldsymbol{P(A{>}B)}$ & \shortstack{\textbf{Reverse}\\\textbf{cases}} & \shortstack{\textbf{Mann--Whitney}\\$\boldsymbol{p}$} \\
\midrule
Llama-2   & Mistral    & 618 & 0.998 & 0.2\% & $10^{-200}$ \\
Llama-2   & Qwen-2.5   & 985 & 0.992 & 0.1\% & $<\!10^{-300}$ \\
Llama-2   & Llama-3.1  & 742 & 0.995 & 0.0\% & $10^{-233}$ \\
Llama-2   & Gemma-2-9B & 984 & 0.996 & 0.1\% & $10^{-304}$ \\
Mistral   & Qwen-2.5   & 368 & 0.995 & 0.0\% & $10^{-113}$ \\
Mistral   & Gemma-2-9B & 368 & 0.989 & 0.0\% & $10^{-108}$ \\
\rowcolor{white}
Llama-3.1 & Qwen-2.5   & 241 & 0.975 & 0.0\% & $10^{-71}$ \\
\bottomrule
\end{tabular}
}
\vspace{-1em}
\end{wraptable}

\paragraph{Selective retrieval.}
\label{sec:retrieval}
The probe identifies queries for external verification but does not supply updated answers. For each model, we rank test queries by a trigger score, check the top 20\% against our Wikidata snapshot, and retain the model's answers for the rest. The drift probe identifies 63\% of outdated answers, compared with 49\% for the hazard, 45\% for first-token entropy, and 29\% for an identically trained correctness probe (Appendix~\ref{app:routing}). The comparison with the correctness probe indicates that the gain depends on the signal being learned. Entropy requires a 29\% budget to match the 76.5\% accuracy achieved by the drift probe at a 20\% budget, corresponding to 31\% fewer lookups with the drift probe. Checked queries receive the correct dated answer, so this simulation evaluates how well each trigger prioritizes queries for verification.

\section{Mechanistic Grounding}
\label{sec:mechanism}

Sections~\ref{sec:orthogonality} and~\ref{sec:consequences} have shown that the model encodes drift separately from correctness and uncertainty, that output-level methods miss it, and that it depends on the model. This section asks how. We locate where the model processes the query year, explain why \textsc{Stale-Recall} and \textsc{Confabulation} look alike at the output, and test a separate steering direction. Our experiments suggest that the model weighs competing holders during retrieval and stores the result in its hidden states, but does not use it to produce the answer.

\begin{wraptable}{r}{0.52\linewidth}
  \centering
    \vspace{-2.2em}

\caption{Activation patching summary. Subscripts show
layers. \emph{Recovery}: clean signal restored at patched
position (1.0: full, 0.0: none) (App.~\ref{app:patching}).}
  \label{tab:patching}
  \vspace{0.6em}
  \scriptsize
  \rowcolors{2}{verylightgray}{white}
  \resizebox{\linewidth}{!}{
  \begin{tabular}{lcccr}
  \toprule
  \textbf{Model} & \shortstack{\textbf{Prompt}\\\textbf{pairs}}
    & \shortstack{\textbf{Peak at}\\\textbf{entity}}
    & \shortstack{\textbf{Peak at}\\\textbf{last token}}
    & \shortstack{\textbf{Last minus}\\\textbf{entity}} \\
  \midrule
  \multicolumn{5}{l}{\emph{Entity-routed} (last minus entity $<0$)} \\
  Llama-2-7B   & 75  & \textbf{1.061}\,\textsubscript{L27} & 0.939\,\textsubscript{L31} & $-0.122$ \\
  Mistral-7B   & 93  & \textbf{0.937}\,\textsubscript{L20} & 0.842\,\textsubscript{L31} & $-0.095$ \\
  \midrule
  \multicolumn{5}{l}{\emph{Last-routed} (last minus entity $>0$)} \\
  Llama-3.1-8B & 78  & 0.012\,\textsubscript{L16} & \textbf{0.967}\,\textsubscript{L31} & $+0.955$ \\
  Qwen-2.5-7B  & 77  & 0.005\,\textsubscript{L21} & \textbf{1.003}\,\textsubscript{L27} & $+0.998$ \\
  Gemma-2-9B   & 133 & 0.009\,\textsubscript{L27} & \textbf{0.968}\,\textsubscript{L41} & $+0.959$ \\
  Gemma-2-2B   & 33  & 0.006\,\textsubscript{L17} & \textbf{0.861}\,\textsubscript{L25} & $+0.855$ \\
  \bottomrule
  \end{tabular}
  }
  \vspace{-1.0em}
  \end{wraptable}

\paragraph{The temporal retrieval circuit.}
\label{sec:patching}
We localize temporal processing using single-position activation patching at each layer~\citep{ heimersheim2024use,meng2022locating}.
For each entity--relation pair we construct a clean prompt that the model
answers correctly and a corrupted prompt whose correct answer is a different holder. The prompts differ only in the year token, which precedes the entity so that causal attention allows the year to influence the entity's representation.  At each (layer, position) we
replace the corrupted hidden state with its clean counterpart and measure
the recovery of the clean answer's logit advantage:
\begin{equation}
\label{eq:recovery}
\textstyle
\text{recovery}(\ell, p) =
  (\delta_{\text{patch}} - \delta_{\text{corr}})\,/\,
  (\delta_{\text{clean}} - \delta_{\text{corr}})
\end{equation}

where $\delta = \text{logit}(t_{\text{clean}}) -
\text{logit}(t_{\text{corr}})$ (protocol in
Appendix~\ref{app:patching}).  Three findings emerge
(Table~\ref{tab:patching}).

Temporal information propagates beyond the year-token position.
Recovery at that position remains at 1.0 through the first half of the network, and then drops below 0.05 at late retrieval layers
(L15/L19 on Llama-2; per-model peaks in Table~\ref{tab:patching}), consistent with temporal information shifting to the entity representation. These interventions support a prior correlational finding~\citep{park2025does}.
Where temporal information accumulates depends on the model's
chat-template structure.  In Llama-3.1, Qwen-2.5, and Gemma,
a 4--6 token role-header preamble (e.g.\
\texttt{<|im\_start|>assistant{\textbackslash}n}) provides a dedicated
prediction position, and temporal information aggregates.
Llama-2 and Mistral lack this preamble, and temporal
information instead flows to the entity position, the pattern documented
for base models~\citep{meng2022locating, choe-etal-2025-autoregressive}.
In both regimes settings, the probe layer selected under controlled evaluation (L28 for Llama-2,
L27 for Llama-3.1) lies within the post-retrieval window,
suggesting that the probe reads the representation after temporal routing.  Section~\ref{sec:steering} shows that a separate steering
direction at these layers shifts probability between competing holders.

\paragraph{Direct logit attribution.}
\label{sec:dla}
\begin{wraptable}{r}{0.43\linewidth}
\centering
\vspace{-1.8em}
\caption{MLP trajectory correlation and peak layer
(\textsc{Stale-Recall} vs.\ \textsc{Confabulation}).
$^\dagger$Gemma L0 peak reflects a known DLA
artifact~\citep{janiak2024adversarial}; correlations
excluding L0 remain ${\geq}\,0.96$
(Appendix~\ref{app:dla_no_l0}).}
\vspace{0.5em}
\label{tab:dla}
\scriptsize
\rowcolors{2}{verylightgray}{white}
\resizebox{\linewidth}{!}{
\begin{tabular}{lccc}
\toprule
\textbf{Model} & $\boldsymbol{r}$ & \shortstack{\textbf{Stale-Recall}\\\textbf{peak}} & \shortstack{\textbf{Confabulation}\\\textbf{peak}} \\
\midrule
Llama-2    & 0.986 & L29 & L29 \\
Mistral    & 0.947 & L28 & L28 \\
Llama-3.1  & 0.883 & L29 & L29 \\
Qwen-2.5   & 0.815 & L23 & L23 \\
Gemma-2-9B & 0.993 & L0$^\dagger$  & L0$^\dagger$  \\
\rowcolor{white}
Gemma-2-2B & 0.997 & L0$^\dagger$  & L0$^\dagger$  \\
\bottomrule
\end{tabular}
}
\vspace{-1.0em}
\end{wraptable}
Section~\ref{sec:consequences} shows that output-level methods miss drift. To see why, we ask whether the model produces outdated and fabricated answers in different ways. We use direct logit attribution (DLA)~\citep{elhage2021mathematical}, which splits the logit of the output token into the contribution of each MLP layer, and we compare these per-layer curves for \textsc{Stale-Recall} and \textsc{Confabulation} answers. The two curves correlate above 0.81 on every model (0.815--0.997, mean 0.94), and both peak at the same layer, for example layer 29 on Llama-2 (Table~\ref{tab:dla}). On the two Gemma models, the shared peak at layer 0 is a known artifact of DLA, and the correlation stays above 0.96 when we remove this layer (Appendix~\ref{app:dla_no_l0}). So the MLP layers push toward the answer in nearly the same pattern, whether the answer is outdated or fabricated, and the output carries little trace of the difference. The provenance probe in Section~\ref{sec:decodability}, however, separates the two answers with an AUROC of 0.89--0.99, so the difference exists in the hidden states but does not reach the output. This helps explain why the output-level methods in Section~\ref{sec:consequences} fail.
\paragraph{Causal steering.}
\label{sec:steering}
The last analysis asks what happens when we change the hidden states directly. We build a steering direction $\hat{\mathbf{d}}$ as the unit difference between the mean hidden states of \textsc{Stale-Recall} and \textsc{Stable-Correct} answers. We compute it within the \texttt{head\_of\_government} relation, the largest relation with enough queries in every state, because the mean difference over all relations is dominated by features of wrong answers in general (Section~\ref{sec:dom_dissociation}). $\hat{\mathbf{d}}$ is not the drift probe $\mathbf{w}_d$. Their mean $|\cos|$ is 0.063 at the steering layers and 0.042 at the cross-validated layers, and they share under 2\% of their variance on every model (Appendix~\ref{app:steering_full}). We test $\hat{\mathbf{d}}$ in two ways. Removing it tests whether the model needs this direction to produce its answer, and adding it tests whether the direction can change the answer. Removing it changes no outputs on Llama-2 across 300 test queries, 100 from each of \textsc{Stale-Recall}, \textsc{Stable-Correct}, and \textsc{Confabulation}, and changes fewer than 10\% of outputs on Mistral and Llama-3.1 (Appendix~\ref{app:steering_full}). So the model does not rely on this direction when it answers. Adding it, scaled to the mean norm of the hidden states, moves probability between holders. On \textsc{Stale-Recall}, the logit of the outdated holder drops on all five models, while the logit of the current holder rises or stays the same (Table~\ref{tab:steering}). On Llama-2, for example, the logit of the outdated holder drops by 3.50, while that of the current holder rises by 0.03. The same shift appears on \textsc{Stable-Correct} and \textsc{Confabulation} (Appendix~\ref{app:steering_full}), so $\hat{\mathbf{d}}$ encodes holder competition, that is, how strongly other candidate answers were active during retrieval, and it is not specific to outdated facts. The direction therefore has no effect when removed but a clear effect when amplified. Makelov et al.~\citep{ICLR2024_70b8505a} show that such interventions can act through pathways the model does not normally use. Here, amplification moves probability between specific competing holders, which suggests that it acts on the temporal retrieval circuit and does not spread broadly. These results describe $\hat{\mathbf{d}}$, while the drift probe reads a different direction, whether a fact has drifted. Together with the patching and DLA results, they support the account from the start of this section, in which the model weighs competing holders during retrieval and stores the result in its hidden states, but does not use it to produce the answer.

\begin{wraptable}{r}{0.42\linewidth}
\centering
\vspace{-1em}
\caption{Logit changes under $+\hat{\mathbf{d}}$ amplification
on \textsc{Stale-Recall}. The last column shows selectivity, which is calculated as Selectivity =
Outdated\,$\Delta$ $-$ Current\,$\Delta$.}
\vspace{1.em}
\label{tab:steering}
\scriptsize
\rowcolors{2}{verylightgray}{white}
\renewcommand{\arraystretch}{1.3}
\resizebox{\linewidth}{!}{
\begin{tabular}{lccc}
\toprule
\textbf{Model} & \shortstack{\textbf{Outdated}\\\textbf{$\Delta$}} & \shortstack{\textbf{Current}\\\textbf{$\Delta$}} & \textbf{Selec.} \\
\midrule
Llama-2   & $-3.50$ & $+0.03$ & $-3.53$ \\
Mistral   & $-3.56$ & $+0.20$ & $-3.76$ \\
Llama-3.1 & $-1.74$ & $+0.72$ & $-2.46$ \\
Qwen-2.5  & $-2.70$ & $+0.53$ & $-3.23$ \\
\rowcolor{white}
Gemma-9B  & $-6.51$ & $+4.04$ & $-10.55$ \\
\bottomrule
\end{tabular}
}
\end{wraptable}
\label{sec:rag}
\begin{figure}[t]
\centering
\includegraphics[width=1.0\textwidth]{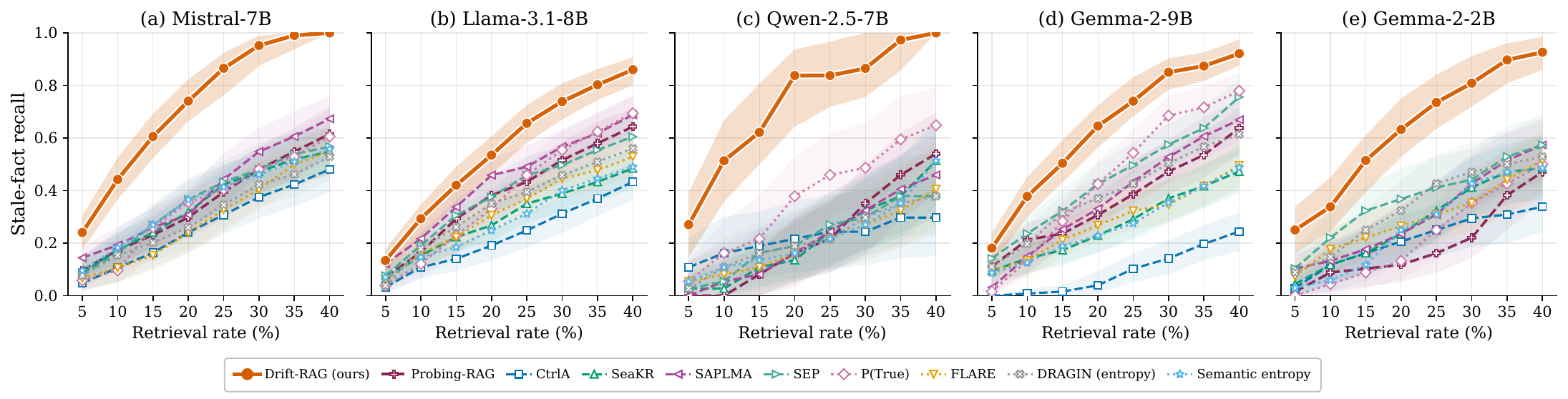}
\caption{Drift-aware selective retrieval across five models. At every retrieval rate,
Drift-RAG (ours) recovers far more stale-fact errors than any adaptive-RAG trigger;
baselines cluster near the random-retrieval reference.}
\label{fig:drift-rag-main}
\end{figure}

\paragraph{Comparison with adaptive retrieval triggers.}
This experiment complements the routing simulation of Section~\ref{sec:retrieval} with a different protocol: it scores queries out-of-fold and counts stale-fact errors on the answerable slice only, so its numbers are not directly comparable to the routing results. Every adaptive retrieval-augmented generation (RAG) method decides when to retrieve from an internal signal, whether output confidence~\citep{jiang2023active,su2024dragin},
self-evaluation~\citep{kadavath2022language}, or a representation-space confidence direction (CtrlA~\citep{huanshuo-etal-2025-ctrla},
SeaKR~\citep{yao-etal-2025-seakr}). Each trigger scores every query, and the top-scoring fraction is sent to retrieval. At an equal 20\% budget, the drift trigger recovers a mean 66\% of stale-fact
errors against at most 34\% for any baseline (SEP), and the gap holds
across the full 5--40\% budget sweep
(Figure~\ref{fig:drift-rag-main}; Appendix~\ref{app:rag}). CtrlA and SeaKR read the residual stream exactly as our probe
does, yet still miss stale facts, because the directions they extract encode
\emph{confidence} rather than temporal validity. This is consistent with our finding that signals of \emph{confidence} and temporal drift capture different information: access to hidden states alone is insufficient unless the retrieval trigger targets temporal validity.

\section{Conclusion and Future Work}
\label{sec:conclusion}
In this work, we showed that the hidden states of six LLMs we test carry a signal for whether a fact has drifted since their training cutoff, even when the model gives the outdated answer confidently. To test this, we built a dataset of 3{,}511 queries over 1{,}342 Wikidata facts with dated timelines. We then trained a linear probe to detect this drift. Across six instruction-tuned models, it reached an AUROC of 0.80--0.93, while prior methods based on correctness, uncertainty, and consistency stayed near chance. We found that the drift probe was almost perpendicular to the correctness and uncertainty probes, so the models encoded drift separately from whether they were right or unsure. To analyze this from the model internals, we performed a mechanistic analysis, which showed that the output-level detectors failed because the MLP layers that retrieve the answer contributed to outdated and fabricated answers in almost the same pattern. Finally, to test whether the drift probe is useful in practice, we used it in selective retrieval methods, where it found more outdated answers than the baselines with the same number of lookups.  

In future work, we plan to extend the analysis to more domains and to other types of knowledge that change over time, such as software APIs, legislation, or medical guidelines. We also plan to test larger models and base models, since here we studied only instruction-tuned models with up to 9B parameters. This would help establish how model scale and instruction tuning affect the representation of temporal drift. Another direction is to examine whether probes generalize to new domains and future time periods with fewer labeled examples, especially for facts with limited temporal evidence in pretraining. Finally, we plan to extend selective retrieval with longer answers that mix current and outdated information, testing whether the drift signal can guide targeted verification and correction.

\bibliographystyle{plain}
\bibliography{custom}

@article{dhingra-etal-2022-time,
    title = "Time-Aware Language Models as Temporal Knowledge Bases",
    author = "Dhingra, Bhuwan  and
      Cole, Jeremy R.  and
      Eisenschlos, Julian Martin  and
      Gillick, Daniel  and
      Eisenstein, Jacob  and
      Cohen, William W.",
    editor = "Roark, Brian  and
      Nenkova, Ani",
    journal = "Transactions of the Association for Computational Linguistics",
    volume = "10",
    year = "2022",
    address = "Cambridge, MA",
    NOpublisher = "MIT Press",
    url = "https://aclanthology.org/2022.tacl-1.15/",
    doi = "10.1162/tacl_a_00459",
    pages = "257--273"
}

@inproceedings{liska2022streamingqa,
  title={{StreamingQA}: {A} Benchmark for Adaptation to New Knowledge over Time in Question Answering Models},
  author={Liska, Adam and Kocisky, Tomas and Gribovskaya, Elena and Terzi, Tayfun and Sezener, Eren and Agrawal, Devang and De Masson D'Autume, Cyprien and Scholtes, Tim and Zaheer, Manzil and Young, Susannah and Gilsenan-Mcmahon, Ellen and Austin, Sophia and Blunsom, Phil and Lazaridou, Angeliki},
  booktitle={Proceedings of the 39th International Conference on Machine Learning},
  pages={13604--13622},
  year={2022},
  
  volume = {162},
  NOpublisher = {PMLR},
  url = {https://proceedings.mlr.press/v162/liska22a.html},
  address = {Baltimore, Maryland, USA},
  series = {ICML~'22},
}

@inproceedings{azaria2023internal,
  title={The Internal State of an {LLM} Knows When It's Lying},
  author={Azaria, Amos and Mitchell, Tom},
  booktitle={Findings of the Association for Computational Linguistics: EMNLP 2023},
  pages={967--976},
  year={2023},
  doi = {10.18653/v1/2023.findings-emnlp.68},
  url = {https://aclanthology.org/2023.findings-emnlp.68/},
  NOpublisher = {Association for Computational Linguistics},
  address = {Singapore},
  series = {EMNLP~'23},
}

@inproceedings{kossen2024semantic,
  title={Semantic Entropy Probes: {Robust} and Cheap Hallucination Detection in {LLMs}},
  author={Han, Jiatong and Kossen, Jannik and Razzak, Muhammed and Schut, Lisa and Malik, Shreshth A. and Gal, Yarin},
  
  year={2024},
  url = {https://openreview.net/forum?id=Zd0XLr6JKn},
  booktitle = {ICML 2024 Workshop on Foundation Models in the Wild},
  series = {FM-Wild~'24},
  address = {Vienna, Austria},
}

@inproceedings{patel2026llm,
  title={Are {LLM} Uncertainty and Correctness Encoded by the Same Features? {A} Functional Dissociation via Sparse Autoencoders},
  author={Patel, Het and Chen, Tiejin and Wei, Hua and Papalexakis, Evangelos E. and Chen, Jia},
  
  year={2026},
  url = {https://arxiv.org/abs/2604.19974},
  booktitle = {Proceedings of the 2026 SIAM International Conference on Data Mining},
  series = {SDM~'26},
  address = {Salt Lake City, Utah, USA},
  note = {Accepted for publication},
}

@inproceedings{marks2023geometry,
  title={The Geometry of Truth: {Emergent} Linear Structure in Large Language Model Representations of True/False Datasets},
  author={Marks, Samuel and Tegmark, Max},
  
  year={2024},
  url = {https://openreview.net/forum?id=aajyHYjjsk},
  booktitle = {First Conference on Language Modeling},
  series = {COLM~'24},
  address = {Philadelphia, PA, USA},
}

@inproceedings{park2023linear,
  title={The Linear Representation Hypothesis and the Geometry of Large Language Models},
  author={Park, Kiho and Choe, Yo Joong and Veitch, Victor},
  
  year={2024},
  volume = {235},
  url = {https://proceedings.mlr.press/v235/park24c.html},
  booktitle = {Proceedings of the 41st International Conference on Machine Learning},
  pages = {39643--39666},
  NOpublisher = {PMLR},
  address = {Vienna, Austria},
  series = {ICML~'24},
}

@inproceedings{meng2022locating,
  title={Locating and Editing Factual Associations in {GPT}},
  author={Meng, Kevin and Bau, David and Andonian, Alex and Belinkov, Yonatan},
  
  volume={35},
  pages={17359--17372},
  year={2022},
  booktitle = {Advances in Neural Information Processing Systems},
  doi = {10.52202/068431-1262},
  url = {https://proceedings.neurips.cc/paper_files/paper/2022/file/6f1d43d5a82a37e89b0665b33bf3a182-Paper-Conference.pdf},
  NOpublisher = {Curran Associates, Inc.},
  address = {New Orleans, Louisiana, USA},
  series = {NeurIPS~'22},
}

@inproceedings{park2025does,
  title={Does Time Have Its Place? {Temporal} Heads: {Where} Language Models Recall Time-specific Information},
  author={Park, Yein and Yoon, Chanwoong and Park, Jungwoo and Jeong, Minbyul and Kang, Jaewoo},
  booktitle={Proceedings of the 63rd Annual Meeting of the Association for Computational Linguistics (Volume 1: Long Papers)},
  pages={16616--16643},
  year={2025},
  doi = {10.18653/v1/2025.acl-long.812},
  url = {https://aclanthology.org/2025.acl-long.812/},
  NOpublisher = {Association for Computational Linguistics},
  address = {Vienna, Austria},
  series = {ACL~'25},
}

@inproceedings{choe-etal-2025-autoregressive,
    title = {Do All Autoregressive Transformers Remember Facts the Same Way? {A} Cross-Architecture Analysis of Recall Mechanisms},
    author = "Choe, Minyeong  and
      Cho, Haehyun  and
      Seo, Changho  and
      Kim, Hyunil",
    editor = "Christodoulopoulos, Christos  and
      Chakraborty, Tanmoy  and
      Rose, Carolyn  and
      Peng, Violet",
    booktitle = "Proceedings of the 2025 Conference on Empirical Methods in Natural Language Processing",
    month = nov,
    year = "2025",
    address = "Suzhou, China",
    NOpublisher = "Association for Computational Linguistics",
    url = "https://aclanthology.org/2025.emnlp-main.1448/",
    doi = "10.18653/v1/2025.emnlp-main.1448",
    pages = "28494--28513",
    ISBN = "979-8-89176-332-6",
  series = {EMNLP~'25},
}

@inproceedings{burns2022discovering,
  title={Discovering Latent Knowledge in Language Models Without Supervision},
  author={Burns, Collin and Ye, Haotian and Klein, Dan and Steinhardt, Jacob},
  
  year={2023},
  url = {https://openreview.net/forum?id=ETKGuby0hcs},
  booktitle = {Proceedings of the International Conference on Learning Representations},
  series = {ICLR~'23},
  address = {Kigali, Rwanda},
}

@article{kadavath2022language,
  title={Language Models (Mostly) Know What They Know},
  author={Kadavath, Saurav and Conerly, Tom and Askell, Amanda and Henighan, Tom and Drain, Dawn and Perez, Ethan and Schiefer, Nicholas and Hatfield-Dodds, Zac and DasSarma, Nova and Tran-Johnson, Eli and Johnston, Scott and El-Showk, Sheer and Jones, Andy and Elhage, Nelson and Hume, Tristan and Chen, Anna and Bai, Yuntao and Bowman, Sam and Fort, Stanislav and Ganguli, Deep and Hernandez, Danny and Jacobson, Josh and Kernion, Jackson and Kravec, Shauna and Lovitt, Liane and Ndousse, Kamal and Olsson, Catherine and Ringer, Sam and Amodei, Dario and Brown, Tom and Clark, Jack and Joseph, Nicholas and Mann, Ben and McCandlish, Sam and Olah, Chris and Kaplan, Jared},
  journal={ArXiv preprint},
  year={2022},
  volume = {arXiv:2207.05221},
  url = {https://arxiv.org/abs/2207.05221},
  eprint = {2207.05221},
  archivePrefix = {arXiv},
}

@inproceedings{belrose2023leace,
  title={{LEACE}: {Perfect} linear concept erasure in closed form},
  author={Belrose, Nora and Schneider-Joseph, David and Ravfogel, Shauli and Cotterell, Ryan and Raff, Edward and Biderman, Stella},
  
  volume={36},
  pages={66044--66063},
  year={2023},
  booktitle = {Advances in Neural Information Processing Systems},
  doi = {10.52202/075280-2884},
  url = {https://proceedings.neurips.cc/paper_files/paper/2023/file/d066d21c619d0a78c5b557fa3291a8f4-Paper-Conference.pdf},
  NOpublisher = {Curran Associates, Inc.},
  address = {New Orleans, Louisiana, USA},
  series = {NeurIPS~'23},
}

@article{huang2025survey,
  title={A Survey on Hallucination in Large Language Models: {Principles}, Taxonomy, Challenges, and Open Questions},
  author={Huang, Lei and Yu, Weijiang and Ma, Weitao and Zhong, Weihong and Feng, Zhangyin and Wang, Haotian and Chen, Qianglong and Peng, Weihua and Feng, Xiaocheng and Qin, Bing and Liu, Ting},
  journal={ACM Transactions on Information Systems},
  volume={43},
  number={2},
  pages={1--55},
  year={2025},
  publisher={Association for Computing Machinery},
  doi = {10.1145/3703155},
  url = {https://doi.org/10.1145/3703155},
}

@inproceedings{cheng2024dated,
  title={Dated Data: {Tracing} Knowledge Cutoffs in Large Language Models},
  author={Cheng, Jeffrey and Marone, Marc and Weller, Orion and Lawrie, Dawn and Khashabi, Daniel and Van Durme, Benjamin},
  
  year={2024},
  url = {https://openreview.net/forum?id=wS7PxDjy6m},
  booktitle = {First Conference on Language Modeling},
  series = {COLM~'24},
  address = {Philadelphia, PA, USA},
}

@inproceedings{ravfogel2020null,
  title={Null It Out: {Guarding} Protected Attributes by Iterative Nullspace Projection},
  author={Ravfogel, Shauli and Elazar, Yanai and Gonen, Hila and Twiton, Michael and Goldberg, Yoav},
  booktitle={Proceedings of the 58th Annual Meeting of the Association for Computational Linguistics},
  pages={7237--7256},
  year={2020},
  doi = {10.18653/v1/2020.acl-main.647},
  url = {https://aclanthology.org/2020.acl-main.647/},
  NOpublisher = {Association for Computational Linguistics},
  address = {Online},
  series = {ACL~'20},
}

@inproceedings{cencerrado2025no,
  title={No Answer Needed: {Predicting} {LLM} Answer Accuracy from Question-Only Linear Probes},
  author={Cencerrado, Iv{\'a}n Vicente Moreno and Masdemont, Arnau Padr{\'e}s and Hawthorne, Anton Gonzalvez and Africa, David Demitri and Pacchiardi, Lorenzo},
  
  year={2026},
  url = {https://openreview.net/forum?id=1QcY6LPcdQ},
  booktitle = {ICLR 2026 Workshop on Principled Design for Trustworthy AI: Interpretability, Robustness, and Safety across Modalities},
  series = {ICLR Workshop~'26},
  address = {Rio de Janeiro, Brazil},
}

@article{miao2026closing,
  title={Closing the Confidence-Faithfulness Gap in Large Language Models},
  author={Miao, Miranda Muqing and Ungar, Lyle},
  journal={ArXiv preprint},
  year={2026},
  volume = {arXiv:2603.25052},
  url = {https://arxiv.org/abs/2603.25052},
  eprint = {2603.25052},
  archivePrefix = {arXiv},
}

@article{cho2026confidence,
  title={The Confidence Manifold: {Geometric} Structure of Correctness Representations in Language Models},
  author={Cho, Seonglae and Wu, Zekun and Da Costa, Kleyton and Koshiyama, Adriano},
  journal={ArXiv preprint},
  year={2026},
  volume = {arXiv:2602.08159},
  url = {https://arxiv.org/abs/2602.08159},
  eprint = {2602.08159},
  archivePrefix = {arXiv},
}

@inproceedings{kuhn2023semantic,
  title={Semantic Uncertainty: {Linguistic} Invariances for Uncertainty Estimation in Natural Language Generation},
  author={Kuhn, Lorenz and Gal, Yarin and Farquhar, Sebastian},
  
  year={2023},
  url = {https://openreview.net/forum?id=VD-AYtP0dve},
  booktitle = {Proceedings of the International Conference on Learning Representations},
  series = {ICLR~'23},
  address = {Kigali, Rwanda},
}

@inproceedings{ICLR2024_70b8505a,
 author = {Makelov, Aleksandar and Lange, Georg and Geiger, Atticus and Nanda, Neel},
 booktitle = {Proceedings of the International Conference on Learning Representations},
 editor = {B. Kim and Y. Yue and S. Chaudhuri and K. Fragkiadaki and M. Khan and Y. Sun},
 pages = {26486--26515},
 title = {Is This the Subspace You Are Looking for? {An} Interpretability Illusion for Subspace Activation Patching},
 url = {https://proceedings.iclr.cc/paper_files/paper/2024/file/70b8505ac79e3e131756f793cd80eb8d-Paper-Conference.pdf},
 
 year = {2024},
  address = {Vienna, Austria},
  series = {ICLR~'24},
}

@inproceedings{janiak2024adversarial,
  title={An Adversarial Example for Direct Logit Attribution: {Memory} Management in {GELU-4L}},
  author={Janiak, Jett and Rager, Can and Dao, James and Lau, Yeu-Tong},
  booktitle={Proceedings of the 7th BlackboxNLP Workshop: Analyzing and Interpreting Neural Networks for NLP},
  pages={232--237},
  year={2024},
  doi = {10.18653/v1/2024.blackboxnlp-1.15},
  url = {https://aclanthology.org/2024.blackboxnlp-1.15/},
  NOpublisher = {Association for Computational Linguistics},
  address = {Miami, Florida, US},
  series = {BlackboxNLP~'24},
}

@article{elhage2021mathematical,
  title={A Mathematical Framework for Transformer Circuits},
  author={Elhage, Nelson and Nanda, Neel and Olsson, Catherine and Henighan, Tom and Joseph, Nicholas and Mann, Ben and Askell, Amanda and Bai, Yuntao and Chen, Anna and Conerly, Tom and DasSarma, Nova and Drain, Dawn and Ganguli, Deep and Hatfield-Dodds, Zac and Hernandez, Danny and Jones, Andy and Kernion, Jackson and Lovitt, Liane and Ndousse, Kamal and Amodei, Dario and Brown, Tom and Clark, Jack and Kaplan, Jared and McCandlish, Sam and Olah, Chris},
  journal={Transformer Circuits Thread},
  
  
  
  year={2021},
  url = {https://transformer-circuits.pub/2021/framework/index.html},
}

@article{farquhar2024detecting,
  title={Detecting hallucinations in large language models using semantic entropy},
  author={Farquhar, Sebastian and Kossen, Jannik and Kuhn, Lorenz and Gal, Yarin},
  journal={Nature},
  volume={630},
  number={8017},
  pages={625--630},
  year={2024},
  publisher={Nature Publishing Group UK London},
  doi = {10.1038/s41586-024-07421-0},
  url = {https://www.nature.com/articles/s41586-024-07421-0},
}

@inproceedings{chen2021dataset,
  title={A Dataset for Answering Time-Sensitive Questions},
  author={Chen, Wenhu and Wang, Xinyi and Wang, William Yang},
  
  year={2021},
  volume = {1},
  url = {https://datasets-benchmarks-proceedings.neurips.cc/paper_files/paper/2021/file/1f0e3dad99908345f7439f8ffabdffc4-Paper-round2.pdf},
  booktitle = {Proceedings of the Neural Information Processing Systems Track on Datasets and Benchmarks},
  address = {Online},
  series = {NeurIPS~'21},
}

@inproceedings{vu2024freshllms,
  title={{FreshLLMs}: {Refreshing} Large Language Models with Search Engine Augmentation},
  author={Vu, Tu and Iyyer, Mohit and Wang, Xuezhi and Constant, Noah and Wei, Jerry and Wei, Jason and Tar, Chris and Sung, Yun-Hsuan and Zhou, Denny and Le, Quoc and Luong, Thang},
  booktitle={Findings of the Association for Computational Linguistics: ACL 2024},
  pages={13697--13720},
  year={2024},
  doi = {10.18653/v1/2024.findings-acl.813},
  url = {https://aclanthology.org/2024.findings-acl.813/},
  NOpublisher = {Association for Computational Linguistics},
  address = {Bangkok, Thailand},
  series = {ACL~'24},
}

@inproceedings{meng2022mass,
  title={Mass-Editing Memory in a Transformer},
  author={Meng, Kevin and Sharma, Arnab Sen and Andonian, Alex and Belinkov, Yonatan and Bau, David},
  
  year={2023},
  url = {https://openreview.net/forum?id=MkbcAHIYgyS},
  booktitle = {Proceedings of the International Conference on Learning Representations},
  series = {ICLR~'23},
  address = {Kigali, Rwanda},
}

@inproceedings{mitchell2022memory,
  title={Memory-Based Model Editing at Scale},
  author={Mitchell, Eric and Lin, Charles and Bosselut, Antoine and Manning, Christopher D and Finn, Chelsea},
  booktitle={Proceedings of the 39th International Conference on Machine Learning},
  pages={15817--15831},
  year={2022},
  
  volume = {162},
  NOpublisher = {PMLR},
  url = {https://proceedings.mlr.press/v162/mitchell22a.html},
  address = {Baltimore, Maryland, USA},
  series = {ICML~'22},
}

@inproceedings{ICLR2024_25f7be96,
 author = {Asai, Akari and Wu, Zeqiu and Wang, Yizhong and Sil, Avi and Hajishirzi, Hannaneh },
 booktitle = {Proceedings of the International Conference on Learning Representations},
 editor = {B. Kim and Y. Yue and S. Chaudhuri and K. Fragkiadaki and M. Khan and Y. Sun},
 pages = {9112--9141},
 title = {{Self-RAG}: {Learning} to Retrieve, Generate, and Critique through Self-Reflection},
 url = {https://proceedings.iclr.cc/paper_files/paper/2024/file/25f7be9694d7b32d5cc670927b8091e1-Paper-Conference.pdf},
 
 year = {2024},
  address = {Vienna, Austria},
  series = {ICLR~'24},
}

@inproceedings{jiang2023active,
  title={Active Retrieval Augmented Generation},
  author={Jiang, Zhengbao and Xu, Frank F and Gao, Luyu and Sun, Zhiqing and Liu, Qian and Dwivedi-Yu, Jane and Yang, Yiming and Callan, Jamie and Neubig, Graham},
  booktitle={Proceedings of the 2023 Conference on Empirical Methods in Natural Language Processing},
  pages={7969--7992},
  year={2023},
  doi = {10.18653/v1/2023.emnlp-main.495},
  url = {https://aclanthology.org/2023.emnlp-main.495/},
  NOpublisher = {Association for Computational Linguistics},
  address = {Singapore},
  series = {EMNLP~'23},
}

@inproceedings{mallen2023not,
  title={When Not to Trust Language Models: {Investigating} Effectiveness of Parametric and Non-Parametric Memories},
  author={Mallen, Alex and Asai, Akari and Zhong, Victor and Das, Rajarshi and Khashabi, Daniel and Hajishirzi, Hannaneh},
  booktitle={Proceedings of the 61st Annual Meeting of the Association for Computational Linguistics (Volume 1: Long Papers)},
  pages={9802--9822},
  year={2023},
  doi = {10.18653/v1/2023.acl-long.546},
  url = {https://aclanthology.org/2023.acl-long.546/},
  NOpublisher = {Association for Computational Linguistics},
  address = {Toronto, Canada},
  series = {ACL~'23},
}

@inproceedings{alain2016understanding,
  title = {Understanding intermediate layers using linear classifier probes},
  author = {Alain, Guillaume and Bengio, Yoshua},
  booktitle = {5th International Conference on Learning Representations (Workshop Track)},
  series = {ICLR Workshop~'17},
  address = {Toulon, France},
  year = {2017},
  url = {https://openreview.net/forum?id=HJ4-rAVtl},
}

@inproceedings{tenney2019bert,
  title={{BERT} Rediscovers the Classical {NLP} Pipeline},
  author={Tenney, Ian and Das, Dipanjan and Pavlick, Ellie},
  booktitle={Proceedings of the 57th Annual Meeting of the Association for Computational Linguistics},
  pages={4593--4601},
  year={2019},
  doi = {10.18653/v1/P19-1452},
  url = {https://aclanthology.org/P19-1452/},
  NOpublisher = {Association for Computational Linguistics},
  address = {Florence, Italy},
  series = {ACL~'19},
}

@inproceedings{wang2022interpretability,
  title={Interpretability in the Wild: a Circuit for Indirect Object Identification in {GPT-2} small},
  author={Wang, Kevin and Variengien, Alexandre and Conmy, Arthur and Shlegeris, Buck and Steinhardt, Jacob},
  
  year={2023},
  url = {https://openreview.net/forum?id=NpsVSN6o4ul},
  booktitle = {Proceedings of the International Conference on Learning Representations},
  series = {ICLR~'23},
  address = {Kigali, Rwanda},
}

@article{belrose2023eliciting,
  title={Eliciting Latent Predictions from Transformers with the Tuned Lens},
  author={Belrose, Nora and Ostrovsky, Igor and McKinney, Lev and Furman, Zach and Smith, Logan and Halawi, Danny and Biderman, Stella and Steinhardt, Jacob},
  journal={ArXiv preprint},
  year={2023},
  volume = {arXiv:2303.08112},
  url = {https://arxiv.org/abs/2303.08112},
  eprint = {2303.08112},
  archivePrefix = {arXiv},
}

@inproceedings{geva2022transformer,
  title={Transformer Feed-Forward Layers Build Predictions by Promoting Concepts in the Vocabulary Space},
  author={Geva, Mor and Caciularu, Avi and Wang, Kevin and Goldberg, Yoav},
  booktitle={Proceedings of the 2022 Conference on Empirical Methods in Natural Language Processing},
  pages={30--45},
  year={2022},
  doi = {10.18653/v1/2022.emnlp-main.3},
  url = {https://aclanthology.org/2022.emnlp-main.3/},
  NOpublisher = {Association for Computational Linguistics},
  address = {Abu Dhabi, United Arab Emirates},
  series = {EMNLP~'22},
}

@article{turner2023steering,
  title={Steering Language Models With Activation Engineering},
  author={Turner, Alexander Matt and Thiergart, Lisa and Leech, Gavin and Udell, David and Vazquez, Juan J and Mini, Ulisse and MacDiarmid, Monte},
  journal={ArXiv preprint},
  year={2023},
  volume = {arXiv:2308.10248},
  url = {https://arxiv.org/abs/2308.10248},
  eprint = {2308.10248},
  archivePrefix = {arXiv},
}

@inproceedings{subramani2022extracting,
  title={Extracting Latent Steering Vectors from Pretrained Language Models},
  author={Subramani, Nishant and Suresh, Nivedita and Peters, Matthew E},
  booktitle={Findings of the Association for Computational Linguistics: ACL 2022},
  pages={566--581},
  year={2022},
  doi = {10.18653/v1/2022.findings-acl.48},
  url = {https://aclanthology.org/2022.findings-acl.48/},
  NOpublisher = {Association for Computational Linguistics},
  address = {Dublin, Ireland},
  series = {ACL~'22},
}

@inproceedings{su2024dragin,
  title={{DRAGIN}: {Dynamic} Retrieval Augmented Generation based on the Real-time Information Needs of Large Language Models},
  author={Su, Weihang and Tang, Yichen and Ai, Qingyao and Wu, Zhijing and Liu, Yiqun},
  booktitle={Proceedings of the 62nd Annual Meeting of the Association for Computational Linguistics (Volume 1: Long Papers)},
  pages={12991--13013},
  year={2024},
  doi = {10.18653/v1/2024.acl-long.702},
  url = {https://aclanthology.org/2024.acl-long.702/},
  NOpublisher = {Association for Computational Linguistics},
  address = {Bangkok, Thailand},
  series = {ACL~'24},
}

@inproceedings{yao-etal-2025-seakr,
    title = {{SeaKR}: {Self-aware} Knowledge Retrieval for Adaptive Retrieval Augmented Generation},
    author = {Yao, Zijun  and
Qi, Weijian  and
Pan, Liangming  and
Cao, Shulin  and
Hu, Linmei  and
Liu, Weichuan  and
Hou, Lei  and
Li, Juanzi},
    editor = "Che, Wanxiang  and
      Nabende, Joyce  and
      Shutova, Ekaterina  and
      Pilehvar, Mohammad Taher",
    booktitle = {Proceedings of the 63rd Annual Meeting of the Association for Computational Linguistics (Volume 1: Long Papers)},
    month = jul,
    year = "2025",
    address = "Vienna, Austria",
    NOpublisher = "Association for Computational Linguistics",
    url = "https://aclanthology.org/2025.acl-long.1312/",
    doi = "10.18653/v1/2025.acl-long.1312",
    pages = "27022--27043",
    ISBN = "979-8-89176-251-0",
  series = {ACL~'25},
}

@inproceedings{huanshuo-etal-2025-ctrla,
    title = {{CtrlA}: {Adaptive} Retrieval-Augmented Generation via Inherent Control},
    author = {Liu, Huanshuo  and
Zhang, Hao  and
Guo, Zhijiang  and
Wang, Jing  and
Dong, Kuicai  and
Li, Xiangyang  and
Lee, Yi Quan  and
Zhang, Cong  and
Liu, Yong},
    editor = "Che, Wanxiang  and
      Nabende, Joyce  and
      Shutova, Ekaterina  and
      Pilehvar, Mohammad Taher",
    booktitle = {Findings of the Association for Computational Linguistics: ACL 2025},
    month = jul,
    year = "2025",
    address = "Vienna, Austria",
    NOpublisher = "Association for Computational Linguistics",
    url = "https://aclanthology.org/2025.findings-acl.652/",
    doi = "10.18653/v1/2025.findings-acl.652",
    pages = "12592--12618",
    ISBN = "979-8-89176-256-5",
  series = {ACL~'25},
}

@inproceedings{baek2025probing,
  title={{Probing-RAG}: {Self-Probing} to Guide Language Models in Selective Document Retrieval},
  author={Baek, Ingeol and Chang, Hwan and Kim, Byeongjeong and Lee, Jimin and Lee, Hwanhee},
  booktitle={Findings of the Association for Computational Linguistics: NAACL 2025},
  pages={3287--3304},
  year={2025},
  doi = {10.18653/v1/2025.findings-naacl.181},
  url = {https://aclanthology.org/2025.findings-naacl.181/},
  NOpublisher = {Association for Computational Linguistics},
  address = {Albuquerque, New Mexico},
  series = {NAACL~'25},
}

@article{heimersheim2024use,
    title = {How to use and interpret activation patching},
    author = {Stefan Heimersheim and Neel Nanda},
    journal = {ArXiv preprint},
    volume = {arXiv:2404.15255},
    year = {2024},
  url = {https://arxiv.org/abs/2404.15255},
  eprint = {2404.15255},
  archivePrefix = {arXiv},
}

@inproceedings{soldaini2024dolma,
  title={Dolma: an Open Corpus of Three Trillion Tokens for Language Model Pretraining Research},
  author={Soldaini, Luca and Kinney, Rodney and Bhagia, Akshita and Schwenk, Dustin and Atkinson, David and Authur, Russell and Bogin, Ben and Chandu, Khyathi and Dumas, Jennifer and Elazar, Yanai and Hofmann, Valentin and Jha, Ananya and Kumar, Sachin and Lucy, Li and Lyu, Xinxi and Lambert, Nathan and Magnusson, Ian and Morrison, Jacob and Muennighoff, Niklas and Naik, Aakanksha and Nam, Crystal and Peters, Matthew and Ravichander, Abhilasha and Richardson, Kyle and Shen, Zejiang and Strubell, Emma and Subramani, Nishant and Tafjord, Oyvind and Walsh, Evan and Zettlemoyer, Luke and Smith, Noah and Hajishirzi, Hannaneh and Beltagy, Iz and Groeneveld, Dirk and Dodge, Jesse and Lo, Kyle},
  booktitle={Proceedings of the 62nd Annual Meeting of the Association for Computational Linguistics (Volume 1: Long Papers)},
  pages={15725--15788},
  year={2024},
  doi = {10.18653/v1/2024.acl-long.840},
  url = {https://aclanthology.org/2024.acl-long.840/},
  NOpublisher = {Association for Computational Linguistics},
  address = {Bangkok, Thailand},
  series = {ACL~'24},
}

@inproceedings{zhang-choi-2023-mitigating,
    title = "Mitigating Temporal Misalignment by Discarding Outdated Facts",
    author = "Zhang, Michael  and
      Choi, Eunsol",
    editor = "Bouamor, Houda  and
      Pino, Juan  and
      Bali, Kalika",
    booktitle = "Proceedings of the 2023 Conference on Empirical Methods in Natural Language Processing",
    month = dec,
    year = "2023",
    address = "Singapore",
    NOpublisher = "Association for Computational Linguistics",
    url = "https://aclanthology.org/2023.emnlp-main.879/",
    doi = "10.18653/v1/2023.emnlp-main.879",
    pages = "14213--14226"
}

@inproceedings{mousavi-etal-2024-dyknow,
    title = "{D}y{K}now: Dynamically Verifying Time-Sensitive Factual Knowledge in {LLM}s",
    author = "Mousavi, Seyed Mahed  and
      Alghisi, Simone  and
      Riccardi, Giuseppe",
    editor = "Al-Onaizan, Yaser  and
      Bansal, Mohit  and
      Chen, Yun-Nung",
    booktitle = "Findings of the {A}ssociation for {C}omputational {L}inguistics: {EMNLP} 2024",
    month = nov,
    year = "2024",
    address = "Miami, Florida, USA",
    NOpublisher = "Association for Computational Linguistics",
    url = "https://aclanthology.org/2024.findings-emnlp.471/",
    doi = "10.18653/v1/2024.findings-emnlp.471",
    pages = "8014--8029"
}

@article{cox1972regression,
  title={Regression models and life-tables},
  author={Cox, David R},
  journal={Journal of the royal statistical society: Series B (methodological)},
  volume={34},
  number={2},
  pages={187--202},
  year={1972},
  publisher={Wiley Online Library}
}

@book{kalbfleisch2002statistical,
  title={The statistical analysis of failure time data},
  author={Kalbfleisch, John D and Prentice, Ross L},
  year={2002},
  publisher={John Wiley \& Sons}
}

@inproceedings{wang-etal-2024-factuality,
    title = "Factuality of Large Language Models: A Survey",
    author = "Wang, Yuxia  and
      Wang, Minghan  and
      Manzoor, Muhammad Arslan  and
      Liu, Fei  and
      Georgiev, Georgi Nenkov  and
      Das, Rocktim Jyoti  and
      Nakov, Preslav",
    editor = "Al-Onaizan, Yaser  and
      Bansal, Mohit  and
      Chen, Yun-Nung",
    booktitle = "Proceedings of the 2024 Conference on Empirical Methods in Natural Language Processing",
    month = nov,
    year = "2024",
    address = "Miami, Florida, USA",
    NOpublisher = "Association for Computational Linguistics",
    url = "https://aclanthology.org/2024.emnlp-main.1088/",
    doi = "10.18653/v1/2024.emnlp-main.1088",
    pages = "19519--19529"
}

@inproceedings{shelmanov-etal-2025-head,
    title = "A Head to Predict and a Head to Question: Pre-trained Uncertainty Quantification Heads for Hallucination Detection in {LLM} Outputs",
    author = "Shelmanov, Artem  and
      Fadeeva, Ekaterina  and
      Tsvigun, Akim  and
      Tsvigun, Ivan  and
      Xie, Zhuohan  and
      Kiselev, Igor  and
      Daheim, Nico  and
      Zhang, Caiqi  and
      Vazhentsev, Artem  and
      Sachan, Mrinmaya  and
      Nakov, Preslav  and
      Baldwin, Timothy",
    editor = "Christodoulopoulos, Christos  and
      Chakraborty, Tanmoy  and
      Rose, Carolyn  and
      Peng, Violet",
    booktitle = "Proceedings of the 2025 Conference on Empirical Methods in Natural Language Processing",
    month = nov,
    year = "2025",
    address = "Suzhou, China",
    NOpublisher = "Association for Computational Linguistics",
    url = "https://aclanthology.org/2025.emnlp-main.1809/",
    doi = "10.18653/v1/2025.emnlp-main.1809",
    pages = "35712--35731",
    ISBN = "979-8-89176-332-6"
}

@inproceedings{xie-etal-2025-fire,
    title = "{FIRE}: Fact-checking with Iterative Retrieval and Verification",
    author = "Xie, Zhuohan  and
      Xing, Rui  and
      Wang, Yuxia  and
      Geng, Jiahui  and
      Iqbal, Hasan  and
      Sahnan, Dhruv  and
      Gurevych, Iryna  and
      Nakov, Preslav",
    editor = "Chiruzzo, Luis  and
      Ritter, Alan  and
      Wang, Lu",
    booktitle = "Findings of the {A}ssociation for {C}omputational {L}inguistics: {NAACL} 2025",
    month = apr,
    year = "2025",
    address = "Albuquerque, New Mexico, USA",
    NOpublisher = "Association for Computational Linguistics",
    url = "https://aclanthology.org/2025.findings-naacl.158/",
    doi = "10.18653/v1/2025.findings-naacl.158",
    pages = "2901--2914",
    ISBN = "979-8-89176-195-7"
}

\appendix
\clearpage
{\huge \textbf{Appendix}}
\section{Dataset Details}
\label{app:dataset}

Table~\ref{tab:models} lists the six we experimented with models, their checkpoints, and the
training cutoffs we use to assign drift labels.
Table~\ref{tab:cell_dist} gives the number of samples in each temporal
knowledge state for each model. The counts vary because each cutoff
leaves a different window of post-cutoff query years and because models
differ in how often they confabulate. Llama-2, which has the earliest
cutoff, has the most \textsc{Stale-Recall} and \textsc{Confabulation}
samples. Appendix~\ref{app:cell_assignment} describes how we assign
each sample to a cell.

\begin{table}[h]
\centering
\caption{Models with HuggingFace checkpoints. All loaded in float16
on a single A100.}
\label{tab:models}
\footnotesize
\setlength{\tabcolsep}{3pt}
\begin{tabular}{llcccl}
\toprule
\bf Model & \bf Family & \bf Parameters & \bf Cutoff & \bf Vocabulary & \bf Checkpoint \\
\midrule
Llama-2-7B-Chat       & Llama   & 7B & Sep 2022 & 32K
  & \texttt{meta-llama/Llama-2-7b-chat-hf} \\
Mistral-7B-Instruct   & Mistral & 7B & Sep 2023 & 32K
  & \texttt{mistralai/Mistral-7B-Instruct-v0.1} \\
Llama-3.1-8B-Instruct & Llama   & 8B & Dec 2023 & 128K
  & \texttt{meta-llama/Llama-3.1-8B-Instruct} \\
Qwen-2.5-7B-Instruct  & Qwen    & 7B & Jun 2024 & 152K
  & \texttt{Qwen/Qwen2.5-7B-Instruct} \\
Gemma-2-9B-IT         & Gemma   & 9B & Jun 2024 & 256K
  & \texttt{google/gemma-2-9b-it} \\
Gemma-2-2B-IT         & Gemma   & 2B & Jun 2024 & 256K
  & \texttt{google/gemma-2-2b-it} \\
\bottomrule
\end{tabular}
\end{table}

\begin{table}[h]
\centering
\caption{Cell counts per model (3{,}511 samples each). Training-only
cells below the midrule are used for class balancing but not analyzed
independently.}
\label{tab:cell_dist}
\footnotesize
\setlength{\tabcolsep}{3pt}
\begin{tabular}{lrrrrrr}
\toprule
& \textbf{Llama-2} & \textbf{Mistral} & \textbf{Llama-3.1}
& \textbf{Qwen-2.5} & \textbf{Gemma-9B} & \textbf{Gemma-2B} \\
& \scriptsize\textit{Sep\,'22} & \scriptsize\textit{Sep\,'23}
& \scriptsize\textit{Dec\,'23} & \scriptsize\textit{Jun\,'24}
& \scriptsize\textit{Jun\,'24} & \scriptsize\textit{Jun\,'24} \\
\midrule
\textsc{Stable-Correct}    & 279 & 422 & 538 & 261 & 763 & 352 \\
\textsc{Anachronism}       & 130 & 279 & 335 & 252 & 545 & 362 \\
\textsc{Stale-Recall}      & 190 &  58 &  76 &  18 &  71 &  41 \\
\textsc{Obsolete-Current}  & 165 &  46 &  81 &  19 &  56 &  29 \\
\textsc{Confabulation}     & 864 & 395 & 332 & 226 & 135 & 187 \\
\midrule
Stable-Confab              & 1{,}611 & 1{,}927 & 2{,}059 & 2{,}682 & 1{,}907 & 2{,}287 \\
Other / excluded           & 272 & 384 & 90 & 53 & 34 & 253 \\
\bottomrule
\end{tabular}
\end{table}
\subsection{Cell-Assignment Procedure}
\label{app:cell_assignment}

Given a sample $(e, r, t, \text{output}, \tau_\theta)$ and the
verified holder timeline, assignment is deterministic.  Outputs
containing template tokens, Unicode control characters, or
${>}50\%$ non-letter content are classified \texttt{EXCLUDED};
refusal phrases trigger a refusal label only when no timeline match
succeeds.  The drift label is set iff the primary holder's tenure
started strictly after $\tau_\theta$.  Outputs are normalized
(NFKD, lowercased, punctuation to space, whitespace collapsed) and
matched against each candidate holder via a five-tier cascade
(first match wins): exact normalized match; significant-token match
(after blacklisting articles, titles, role nouns); last-name match;
single-token substring match; geographic single-token match.
Ambiguous matches default to no-match.  The cell decision then
proceeds:

\begin{itemize}
\item Matched primary holder $\Rightarrow$
      \textsc{Stable-Correct} / \textsc{Drift-Correct}.
\item Matched holder's tenure ended before $\tau_\theta$
      $\Rightarrow$ \textsc{Stale-Recall} (drifted) /
      \textsc{Anachronism} (not drifted).
\item Matched holder in office at $\tau_\theta$, since replaced
      $\Rightarrow$ \textsc{Obsolete-Current} (drifted) /
      \textsc{Stable-Confab} (not drifted).
\item No match $\Rightarrow$ \textsc{Confabulation}.
\end{itemize}

Holders with non-contiguous tenures (comebacks) are resolved to the
most-recently-ended tenure satisfying
$\text{start} \leq \tau_\theta$.  The procedure is covered by 73
regression tests.

\subsection{18-Relation Extension}

\label{app:relations}

To test whether the drift probe generalizes beyond the four relations of the main benchmark, we build an extension with 18 relations. The 14 new relations come from six relation families and add a third answer type: besides persons and organizations, the correct answer can now be a title (Table~\ref{tab:relation_families}). We build them with the same frozen Wikidata snapshot, the same holder timelines, and the same labeling code as the main benchmark, which the 73 regression tests already cover (Appendix~\ref{app:cell_assignment}). Only the data changes, not the method. All results in the main text use the four core relations unless stated otherwise.

\begin{table}[h]
\centering
\caption{Relation families in the 18-relation extension.}
\label{tab:relation_families}
\small
\begin{tabular}{lcl}
\toprule
\bf Relation Family & \bf Relations & \bf Answer Types \\
\midrule
Governance and politics        & 4 & Person, organization \\
Honors and rank                & 2 & Title \\
Sports                         & 2 & Person, organization \\
Corporate leadership           & 3 & Person, organization \\
Corporate structure            & 4 & Organization \\
Media and personal affiliation & 3 & Person, organization \\
\midrule
Total                          & 18 & Person, organization, title \\
\bottomrule
\end{tabular}
\end{table}

We run three evaluations (Table~\ref{tab:relation_results}). We report \emph{pooled} AUROC, which scores all test queries of a model together, so larger relations count more, and \emph{macro} AUROC, which scores each relation separately and then averages, so every relation counts equally. As in the main benchmark, each fact and all of its query years go entirely into training or entirely into testing.

\emph{All 18 relations.} On fact-disjoint splits over the full extension, the probe reaches a mean of 0.918 pooled and 0.882 macro AUROC, so it stays strong when small and large relations count equally.

\emph{Leave one relation out.} For each relation, we train on the other 17, choose the layer and hyperparameters on those 17 only, freeze the probe, and test it on the held-out relation. Mean macro AUROC drops from 0.882 to 0.821, and every model stays well above chance. Moving to an unseen relation is harder than moving to unseen facts within known relations, but the signal does not collapse.

\emph{Trained on four relations.} We train only on the four core relations, choose the layer and hyperparameters on them, freeze the probe, and test it on the 14 new relations. Against the leave-one-out probe evaluated on the same 14 relations, removing three quarters of the training relations costs 0.022 on average (0.785 vs.\ 0.807), and between 0.017 and 0.028 on every model. These results measure transfer to the 14 new relations under two training regimes; they do not isolate which similarities between relations enable that transfer.

\begin{table}[h]
\centering
\caption{Drift probe AUROC on the 18-relation extension. Pooled scores all test queries together; macro averages over relations. The last two columns test on the 14 new relations only.}
\label{tab:relation_results}
\small
\setlength{\tabcolsep}{5pt}
\begin{tabular}{lcccccc}
\toprule
 & \multicolumn{2}{c}{\bf All 18 Relations}
 & \multicolumn{2}{c}{\bf Leave One Relation Out}
 & \multicolumn{2}{c}{\bf 14 New Relations} \\
\cmidrule(lr){2-3} \cmidrule(lr){4-5} \cmidrule(lr){6-7}
\bf Model & \bf Pooled & \bf Macro & \bf Pooled & \bf Macro & \bf Trained on 4 & \bf Trained on 17 \\
\midrule
Llama-2   & 0.925 & 0.918 & 0.865 & 0.845 & 0.814 & 0.832 \\
Mistral   & 0.969 & 0.897 & 0.888 & 0.834 & 0.794 & 0.820 \\
Llama-3.1 & 0.980 & 0.921 & 0.918 & 0.856 & 0.822 & 0.843 \\
Qwen-2.5  & 0.856 & 0.837 & 0.823 & 0.782 & 0.740 & 0.768 \\
Gemma-9B  & 0.946 & 0.948 & 0.889 & 0.875 & 0.838 & 0.861 \\
Gemma-2B  & 0.834 & 0.771 & 0.785 & 0.731 & 0.701 & 0.718 \\
\midrule
Mean      & 0.918 & 0.882 & 0.861 & 0.821 & 0.785 & 0.807 \\
\bottomrule
\end{tabular}
\end{table}

Held-out AUROC varies within relation families: governance and
politics spans 0.790--0.890, corporate structure 0.760--0.850, and
corporate leadership 0.780--0.860. These ranges show variation within
each family, but do not quantify the proportion of variation
explained by family membership. The results support transfer across
the tested relations; generalization to other domains and relation
structures remains open.

\section{Probing and Geometric Independence}
\label{app:probing}
\subsection{Per-Model Layer Shifts Under Calendar Control}
\label{app:layer_shifts}

Table~\ref{tab:layer_shifts_detail} reports the unconstrained and
controlled layers for all six models. On five of six, the controlled
layer differs from the unconstrained one.

\begin{table}[h]
\centering
\caption{Selected layers and AUROC under unconstrained and controlled
evaluation. Layer shift: controlled layer minus unconstrained layer.}
\label{tab:layer_shifts_detail}
\footnotesize
\begin{tabular}{lccccc}
\toprule
 & \multicolumn{2}{c}{\bf Unconstrained} & \multicolumn{2}{c}{\bf Controlled} & \\
\cmidrule(lr){2-3} \cmidrule(lr){4-5}
\bf Model & \bf Layer & \bf AUROC & \bf Layer & \bf AUROC & \bf Layer Shift \\
\midrule
Llama-2    & L30 & 0.975 & L28 & 0.914 & $-2$  \\
Mistral    & L6  & 0.973 & L6  & 0.925 & $0$   \\
Llama-3.1  & L18 & 0.959 & L27 & 0.923 & $+9$  \\
Qwen-2.5   & L8  & 0.961 & L22 & 0.827 & $+14$ \\
Gemma-9B   & L10 & 0.961 & L28 & 0.801 & $+18$ \\
Gemma-2B   & L10 & 0.942 & L16 & 0.828 & $+6$  \\
\bottomrule
\end{tabular}
\end{table}

Llama-2 is the only model where the controlled layer is
\emph{shallower} than the unconstrained one (L30$\to$L28).
Llama-3.1 shifts in the opposite direction (L18$\to$L27), and
Gemma-2-9B exhibits the largest gap (18 layers).  The diversity of
shift directions rules out a single confound mechanism.

\subsection{Anachronism Falsifier: Drift Probe Scores by Cell}
\label{app:anachronism_falsifier}

If the drift probe detected past-holder retrieval rather than
temporal validity, it would assign similar scores to
\textsc{Stale-Recall} and \textsc{Anachronism}.
Table~\ref{tab:per_cell_scores} reports mean drift probe scores
(logit scale) by cell on the controlled split.

\begin{table}[h]
\centering
\caption{Drift probe score (logit) by cell on the controlled split.
\textsc{Anachronism} and \textsc{Stale-Recall} produce near-identical outputs; the probe assigns negative scores to the former
and positive scores to the latter on every model. Gap: \textsc{Stale-Recall} minus \textsc{Anachronism} score.}
\label{tab:per_cell_scores}
\footnotesize
\begin{tabular}{lccccc}
\toprule
\bf Model & \bf Stable-Correct & \bf Anachronism & \bf Stale-Recall & \bf Gap & \bf Mann--Whitney $p$ \\
\midrule
Llama-2    & $-6.28$ & $-5.66$ & $+4.09$ & $9.75$ & $<10^{-3}$ \\
Mistral    & $-3.88$ & $-2.35$ & $+2.81$ & $5.16$ & $<10^{-3}$ \\
Llama-3.1  & $-4.36$ & $-2.61$ & $+2.22$ & $4.83$ & $<10^{-3}$ \\
Qwen-2.5   & $-4.77$ & $-3.21$ & $+0.99$ & $4.20$ & $<0.05$    \\
Gemma-9B   & $-3.22$ & $-2.14$ & $+2.24$ & $4.38$ & $<10^{-3}$ \\
Gemma-2B   & $-3.09$ & $-2.79$ & $+2.18$ & $4.96$ & $<10^{-3}$ \\
\bottomrule
\end{tabular}
\end{table}

The gap ranges from 4.20 to 9.75 logits, with Mann--Whitney $U$
confirming separation at $p < 10^{-3}$ on five of six models
($p < 0.05$ on Qwen-2.5, which has the narrowest post-cutoff
window).

\subsection{Correctness--Uncertainty Score Correlations}
\label{app:cu_correlations}

Table~\ref{tab:cu_score_corr} reports Pearson $r$ between
correctness and uncertainty probe scores on the controlled test set.
The correlation ($r = -0.13$ to $-0.40$) is substantially larger
than any drift-involving pair
(Table~\ref{tab:orthogonality_main}), consistent with the expected
relationship between confidence and accuracy.

\begin{table}[h]
\centering
\caption{Correctness--uncertainty score correlations (Pearson $r$)
on the controlled test set; $\pm$: half-width of the 1{,}000-resample
bootstrap 95\% CI.}
\label{tab:cu_score_corr}
\footnotesize
\begin{tabular}{lc}
\toprule
\bf Model & \bf Correctness--Uncertainty $r$ \\
\midrule
Llama-2   & $-0.13\pm0.08$ \\
Mistral   & $-0.24\pm0.08$ \\
Llama-3.1 & $-0.40\pm0.07$ \\
Qwen-2.5  & $-0.19\pm0.08$ \\
Gemma-9B  & $-0.28\pm0.08$ \\
Gemma-2B  & $-0.22\pm0.09$ \\
\bottomrule
\end{tabular}
\end{table}
\subsection{Null-Space Random-Direction Baselines}
\label{app:nullspace}

To confirm that the null-space projection results
(\S\ref{sec:nullspace}) are not vacuously small, we compare against
random-direction projections.  For each model, we remove two random
unit vectors from the residual stream and retrain each probe,
repeating 100 times.  The P95 of $|\Delta\text{AUROC}|$ under
random removal is 0.0002--0.0006.  The real null-space $|\Delta|$
for drift falls within this random range on 3/6 models and exceeds
it by at most 0.001 on the remaining three, confirming that the
real nuisance directions do not degrade drift more than noise does.

\subsection{DoM vs.\ Trained-Probe Dissociation}
\label{app:dom_dissociation}

\begin{table}[h]
\centering
\caption{\textbf{DoM vs.\ trained-probe cosines.}  DoM directions
for correctness and uncertainty are highly anti-correlated; trained
probes are near-orthogonal.  Drift is orthogonal to both under
either method.}
\label{tab:dom_vs_probe}
\footnotesize
\setlength{\tabcolsep}{2.5pt}
\renewcommand{\arraystretch}{1.15}
\begin{tabular}{@{}lcccccc@{}}
\toprule
 & \multicolumn{3}{c}{\bf $\cos(\text{DoM})$}
 & \multicolumn{3}{c}{\bf $\cos(\text{Probe})$} \\
\cmidrule(lr){2-4} \cmidrule(lr){5-7}
 & $d{\times}c$ & $d{\times}u$ & $c{\times}u$
 & $d{\times}c$ & $d{\times}u$ & $c{\times}u$ \\
\midrule
Llama-2   & $-.27$ & $+.60$ & $-.75$ & $+.01$ & $-.02$ & $+.09$ \\
Mistral   & $-.45$ & $+.09$ & $-.37$ & $+.05$ & $-.00$ & $+.06$ \\
Llama-3.1 & $-.39$ & $+.35$ & $-.92$ & $+.04$ & $-.01$ & $+.15$ \\
Qwen-2.5  & $-.15$ & $+.30$ & $-.74$ & $+.08$ & $-.00$ & $+.05$ \\
Gemma-9B  & $-.23$ & $+.29$ & $-.87$ & $+.01$ & $-.00$ & $+.10$ \\
Gemma-2B  & $-.20$ & $+.52$ & $-.79$ & $+.03$ & $-.01$ & $+.13$ \\
\bottomrule
\end{tabular}
\end{table}

The dissociation between DoM and trained-probe cosines is the
expected geometry when confounded features dominate the mean
shift~\citep{patel2026llm, cho2026confidence}: a feature that
activates on unfamiliar entities shifts both incorrectness and
uncertainty centroids (creating anti-correlation in DoM space)
while contributing nothing to a probe trained to isolate a single
target.  \citep{cencerrado2025no} independently observe that DoM
probes for correctness act as latent confidence axes, consistent
with this conflation.  The regularized probe suppresses such
confounded features and recovers the pure task-specific
direction~\citep{patel2026llm, miao2026closing}.

Drift extends the two-axis functional dissociation reported by
\citep{patel2026llm} via sparse autoencoders to a third axis:
temporal validity is not a component of either correctness or
uncertainty, but a distinct representational property factored
from both.

\section{Baselines, Robustness, and Selective Retrieval}
\label{app:consequences}

\subsection{Volatility-Hazard Baseline}
\label{app:hazard}

The hazard baseline scores drift without any model internals. For each fact and model, we compute the pre-cutoff transition rate $\rho$ as the number of holder changes before $\tau_\theta$ divided by the years observed, and score a query as $s = 1 - \exp[-\rho\,(t_q - \tau_\theta)]$. No activations, generated answers, post-cutoff transitions, or test labels enter the score. A fact with no holder change before the cutoff gets $\rho = 0$ and the same score at every query year, so the hazard cannot rank it. We therefore evaluate both methods on the facts with at least one earlier change. Eligibility is computed per model, since each model's history window ends at its own cutoff (Table~\ref{tab:hazard_eligibility}). Across models, 92\% of queries and 92\% of facts are eligible.

\begin{table}[h]
\centering
\caption{Hazard eligibility per model over the full evaluation set. Earlier cutoffs leave more query years after the cutoff, so Llama-2 has the most queries.}
\label{tab:hazard_eligibility}
\small
\begin{tabular}{lrrrrrr}
\toprule
 & \multicolumn{3}{c}{\bf Queries} & \multicolumn{3}{c}{\bf Facts} \\
\cmidrule(lr){2-4} \cmidrule(lr){5-7}
\bf Model & \bf Total & \bf Eligible & \bf Share (\%) & \bf Total & \bf Eligible & \bf Share (\%) \\
\midrule
Llama-2   & 1{,}994 & 1{,}822 & 91 & 817 & 750 & 92 \\
Mistral   & 1{,}620 & 1{,}492 & 92 & 786 & 727 & 92 \\
Llama-3.1 & 1{,}620 & 1{,}492 & 92 & 786 & 727 & 92 \\
Qwen-2.5  & 1{,}136 & 1{,}054 & 93 & 728 & 676 & 93 \\
Gemma-9B  & 1{,}136 & 1{,}054 & 93 & 728 & 676 & 93 \\
Gemma-2B  & 1{,}136 & 1{,}054 & 93 & 728 & 676 & 93 \\
\bottomrule
\end{tabular}
\end{table}

We compute AUROC within each model and then average over models (Table~\ref{tab:hazard}). Relation-pooled AUROC scores all eligible relations together within a model; macro-relation AUROC scores each relation separately and then averages. Both methods run on exactly the same held-out rows, and the probe uses the cross-validated layers. The hazard beats every non-drift method in Table~\ref{tab:baselines}, which confirms that pre-cutoff volatility carries real information about drift. The drift probe still leads it by 0.196 pooled and 0.139 macro. The probe's 0.905 here is above its 0.870 full-split mean because the eligible subset differs from the full split, not because the protocol differs.

Table~\ref{tab:hazard} gives the macro-relation result for each relation. The drift probe leads the hazard on all four relations, by 0.110 on \texttt{head\_coach} to 0.156 on \texttt{owned\_by}.

\begin{table}[h]
\centering
\caption{AUROC of the volatility hazard and the drift probe on the hazard-eligible queries, averaged over the six models. The first four rows score each relation separately. Relation-pooled scores all eligible relations together within a model.}
\label{tab:hazard}
\begin{tabular}{lccc}
\toprule
Relation & Volatility Hazard & Drift Probe & Difference \\
\midrule
\texttt{head\_of\_government} & 0.790 & 0.930 & +0.140 \\
\texttt{head\_coach}          & 0.810 & 0.920 & +0.110 \\
\texttt{chair\_of}            & 0.720 & 0.870 & +0.150 \\
\texttt{owned\_by}            & 0.652 & 0.808 & +0.156 \\
\midrule
Macro-relation average        & 0.743 & 0.882 & +0.139 \\
Relation-pooled               & 0.709 & 0.905 & +0.196 \\
\bottomrule
\end{tabular}
\end{table}


\subsection{Why Consistency-Based Methods Fail on Stale Recall}
\label{app:consistency_failure}

CCS's partial success (0.54 mean AUROC) reflects detection of
internal conflict: on drifted facts, the model encodes both the
stored (stale) answer and partial evidence of the current answer,
producing inconsistency that partly aligns with the CCS direction.
However, non-drifted confabulations produce similar inconsistency,
diluting the signal.  Semantic Entropy faces the opposite problem:
\textsc{Stale-Recall} outputs are perfectly self-consistent the
model retrieves the same outdated answer across independent
samples so consistency-based methods rate them as reliable.
Neither internal conflict nor output consistency isolates temporal
drift.

\subsection{Architecture-Matched Control}
\label{app:architecture_control}

SAPLMA trained directly on \textsc{is\_drifted} (same architecture,
same hidden states, but without controlled layer selection) achieves
0.81 mean AUROC (Table~\ref{tab:baselines}, ``MLP probe (drift)''
row), confirming that the drift signal is accessible to any linear
method.  Our controlled protocol recovers it more completely on five
of six models (0.89 vs.\ 0.81 mean), isolating the contribution of
controlled layer selection and post-cutoff filtering from that of
the drift label itself.

\subsection{Entropy Screening Across Thresholds}
\label{app:entropy_sweep}

The 80th-percentile threshold used in \S\ref{sec:consequences} is
one operating point on a continuum.  The asymmetric failure pattern
holds across the full range: at stricter thresholds (lower
percentiles), the \textsc{Stale-Recall} miss rate increases further
because stale recalls have lower entropy than confabulations; at
more permissive thresholds, \textsc{Stable-Correct} false alarm
rates rise while the miss rate remains substantial.  No single
threshold resolves the asymmetry because \textsc{Stale-Recall}
entropy overlaps with \textsc{Stable-Correct} rather than with
\textsc{Confabulation}.

\subsection{Per-Model Entropy Screening Failure Rates}
\label{app:asymmetric}
\begin{table}
\centering
\scriptsize
\caption{Entropy screening fails asymmetrically. Overconf.: more confident than median \textsc{Stable-Correct}.}
\vspace{0.5em}
\label{tab:asymmetric}
\scriptsize
\rowcolors{2}{verylightgray}{white}
\begin{tabular}{lcc}
\toprule
\textbf{Model} & \textbf{Miss} & \textbf{Overconf.} \\
\midrule
Llama-2   & 60.2\% & 31.7\% \\
Mistral   & 39.7\% & 34.5\% \\
Llama-3.1 & 37.3\% & 30.1\% \\
Qwen-2.5  & 72.2\% & 35.2\% \\
Gemma-9B  & 52.9\% & 25.4\% \\
Gemma-2B  & 67.6\% & 28.3\% \\
\midrule
\rowcolor{white}
\textbf{Mean} & \textbf{55.0\%} & \textbf{30.9\%} \\
\bottomrule
\end{tabular}
\vspace{1.0em}
\end{table}

\subsection{Cross-Cutoff Pair Independence}
\label{app:pair_independence}

Five of the twelve cross-cutoff pairs in Table~\ref{tab:hero}
share Llama-2 as Model~A; the twelve pairs are therefore not
statistically independent.  To verify that the result is not
driven by Llama-2-specific properties, we examine the seven
Llama-2-free pairs separately (Table~\ref{tab:hero_independent}).

\begin{table}[h]
\centering
\caption{Cross-cutoff results on Llama-2-free pairs only. Model A has the earlier cutoff.}
\label{tab:hero_independent}
\footnotesize
\begin{tabular}{llccc}
\toprule
\bf Model A & \bf Model B & \bf Matched Queries & \bf $P(A{>}B)$ & \bf Mann--Whitney $p$ \\
\midrule
Mistral   & Llama-3.1 & 128 & 0.992 & $10^{-39}$  \\
Mistral   & Qwen-2.5  & 368 & 0.995 & $10^{-113}$ \\
Mistral   & Gemma-9B  & 368 & 0.989 & $10^{-108}$ \\
Mistral   & Gemma-2B  & 361 & 0.989 & $10^{-110}$ \\
Llama-3.1 & Qwen-2.5  & 241 & 0.975 & $10^{-71}$  \\
Llama-3.1 & Gemma-9B  & 240 & 0.983 & $10^{-61}$  \\
Llama-3.1 & Gemma-2B  & 236 & 0.979 & $10^{-62}$  \\
\bottomrule
\end{tabular}
\end{table}

All seven Llama-2-free pairs satisfy $P(A{>}B) \geq 0.975$ with
$p \leq 10^{-39}$ on the smallest pair ($n=128$, the 3-month
Mistral$\times$Llama-3.1 cutoff gap).  The result is robust to
the dependence structure.

\subsection{Prompt Robustness}
\label{app:prompt_robustness}

The cross-cutoff experiment keeps the prompt fixed and changes the model. The two tests here keep the model fixed and change the prompt.

\emph{Prompt formats.} We train the probe on the original question format (T0) only, then freeze its weights, layer, and regularization strength, and score the same held-out facts written in four unseen formats (Table~\ref{tab:prompt_formats}). The model, fact, query year, drift label, and split stay the same. Across the 16 cases with an unseen format, the mean change from T0 is 0.034 and the largest is 0.063. The changes do not all go one way: Llama-2 and Mistral drop slightly on average, while Llama-3.1 and Gemma-2B rise slightly. We test English rewordings and noisy input only, not other languages or long-form answers.

\begin{table}[h]
\centering
\caption{Frozen drift probe AUROC across prompt formats. The probe is trained on the original format only. Original: ``Who governs Pennsylvania in 2024?''; year first, possessive: ``In 2024, who was Pennsylvania's governor?''; year first, \emph{of} phrasing: ``In 2024, who was the governor of Pennsylvania?''; indirect: an indirect construction; corrupted: the original format with corrupted characters.}
\label{tab:prompt_formats}
\small
\begin{tabular}{lccccc}
\toprule
\bf Model & \bf Original & \shortstack{\bf Year First,\\\bf Possessive} & \shortstack{\bf Year First,\\\bf \emph{of} Phrasing} & \bf Indirect & \bf Corrupted \\
\midrule
Llama-2    & 0.951 & 0.899 & 0.932 & 0.918 & 0.888 \\
Mistral    & 0.929 & 0.897 & 0.904 & 0.881 & 0.902 \\
Llama-3.1  & 0.932 & 0.953 & 0.941 & 0.964 & 0.949 \\
Gemma-2B   & 0.844 & 0.864 & 0.901 & 0.796 & 0.887 \\
\midrule
Mean       & 0.914 & 0.903 & 0.920 & 0.890 & 0.907 \\
\bottomrule
\end{tabular}
\end{table}

\emph{Temporal-cue ablation.} On held-out facts whose prompts contain temporal markers, we strip those markers and rescore with the frozen probe, keeping the fact, entity, relation, query year, label, split, layer, and regularization fixed (Table~\ref{tab:cue_ablation}). Mean AUROC drops by 0.037, from 0.919 to 0.881. Measured against chance, the ablated version keeps 91\% of the above-chance margin, $(0.881 - 0.5)/(0.919 - 0.5)$, so the probe does not run mainly on surface cues in the prompt. The cue-present mean is above the full-split mean because this is a smaller paired subset. This tests robustness to removing cues from the input. It is not a pretraining ablation and cannot show that a fact had no temporal evidence in the training data.

\begin{table}[h]
\centering
\caption{Drift probe AUROC with and without temporal markers in the prompt.}
\label{tab:cue_ablation}
\small
\begin{tabular}{lcc}
\toprule
\bf Model & \bf With Temporal Markers & \bf Without Temporal Markers \\
\midrule
Llama-2    & 0.944 & 0.899 \\
Mistral    & 0.929 & 0.905 \\
Llama-3.1  & 0.960 & 0.925 \\
Gemma-9B   & 0.926 & 0.896 \\
Gemma-2B   & 0.834 & 0.782 \\
\midrule
Mean       & 0.919 & 0.881 \\
\bottomrule
\end{tabular}
\end{table}


\subsection{Coverage}
\label{app:coverage}

\emph{History eligibility.} Ineligible facts are rare, so a single 20\% test split leaves too few to measure. We therefore use fact-grouped 5-fold cross-validation over the evaluation set, with the layer and regularization chosen on the training folds only, so every fact is held out once. The pooled out-of-fold mean is 0.868, which reproduces the 0.870 of Table~\ref{tab:drift_probes}. The probe reaches 0.905 on eligible facts and 0.777 on ineligible ones, and stays above chance on all six models (Table~\ref{tab:coverage_eligibility}). The hazard scores 0.500 on ineligible facts by construction, so the 8\% of facts where the probe is weakest are the same 8\% that the hazard cannot rank. The probe therefore retains useful discrimination even when the history-based hazard provides no ranking information. Pooled AUROC also compares eligible against ineligible facts, so 0.870 sits below the weighted average of 0.905 and 0.777 rather than on it.

\begin{table}[h]
\centering
\vspace{6pt}
\caption{AUROC stratified by history eligibility. Ineligible facts have no recorded holder change before the cutoff. Estimates are less precise on the models with 82 ineligible queries.}
\label{tab:coverage_eligibility}
\small
\begin{tabular}{lcccc}
\toprule
 & & \multicolumn{2}{c}{\bf Drift Probe} & \bf Hazard \\
\cmidrule(lr){3-4} \cmidrule(lr){5-5}
\bf Model & \bf Ineligible Queries & \bf Eligible & \bf Ineligible & \bf Ineligible \\
\midrule
Llama-2   & 172 & 0.947 & 0.812 & 0.500 \\
Mistral   & 128 & 0.953 & 0.806 & 0.500 \\
Llama-3.1 & 128 & 0.953 & 0.810 & 0.500 \\
Qwen-2.5  & 82  & 0.866 & 0.742 & 0.500 \\
Gemma-9B  & 82  & 0.843 & 0.728 & 0.500 \\
Gemma-2B  & 82  & 0.868 & 0.764 & 0.500 \\
\midrule
Mean      &     & 0.905 & 0.777 & 0.500 \\
\bottomrule
\end{tabular}
\end{table}

\emph{Temporal-mention density.} Without the pretraining corpora of these models, we cannot build a truly temporally silent set, so we measure the closest proxy. For each fact, we count how often its holders appear within a 32-token window of a temporal expression (a four-digit year, or \emph{since}, \emph{until}, \emph{former}, \emph{current}, \emph{as of}) in Dolma v1.7~\citep{soldaini2024dolma}, split the test facts into quartiles by that count (boundaries 41, 187, and 903), and run the frozen probe on each quartile (Table~\ref{tab:coverage_density}). AUROC falls with density but does not collapse: the lowest quartile, the closest we can get to temporally silent facts, reaches 0.782. The quartile mean of 0.880 sits above the pooled 0.870 because pooling also compares facts across quartiles. This partition overlaps with eligibility without matching it: 58\% of ineligible facts fall in the lowest quartile. Dolma is a proxy and not the training data of any of these models, so genuinely silent facts remain an open case.

\begin{table}[h]
\centering
\vspace{6pt}
\caption{Drift probe AUROC by quartile of temporal-mention density in Dolma.}
\label{tab:coverage_density}
\small
\begin{tabular}{lcc}
\toprule
\bf Temporal-Mention Density & \bf Share of Test Facts & \bf Drift Probe AUROC \\
\midrule
First quartile (lowest)  & 25\% & 0.782 \\
Second quartile          & 25\% & 0.868 \\
Third quartile           & 25\% & 0.923 \\
Fourth quartile (highest) & 25\% & 0.945 \\
\midrule
Mean                     & 100\% & 0.880 \\
\bottomrule
\end{tabular}
\end{table}

\emph{Cutoff uncertainty.} A model's true cutoff is never known to the day, so we treat it as a range around the stated date. We keep a training fact only if its label is the same for every candidate cutoff in that range, and discard it otherwise. All settings use the same fixed test set on the 18-relation extension (Appendix~\ref{app:relations}), so only the training labels change. With the cutoff known only to within $\pm$12 months, 70.9\% of training facts keep an unambiguous label and the probe reaches 0.840 macro-relation AUROC, against 0.882 at the exact cutoff (Table~\ref{tab:coverage_cutoff}). The method needs a rough bracket around the cutoff, not its exact date.

\begin{table}[h]
\centering
\caption{Drift probe macro-relation AUROC under cutoff uncertainty on the 18-relation extension.}
\label{tab:coverage_cutoff}
\small
\begin{tabular}{lcc}
\toprule
\bf Cutoff Assumption & \bf Training Facts Kept & \bf Macro-Relation AUROC \\
\midrule
Exact stated cutoff        & 100\%  & 0.882 \\
Within $\pm$3 months       & 93.3\% & 0.856 \\
Within $\pm$6 months       & 84.0\% & 0.853 \\
Within $\pm$12 months      & 70.9\% & 0.840 \\
\bottomrule
\end{tabular}
\end{table}

\subsection{Selective Verification}
\label{app:routing}
We simulate selective verification on the fixed fact-disjoint test set. Within each model, we rank the test queries by a trigger score, send the top slice to a dated knowledge base, and keep the model's own answer for the rest. We then average across models. Routed queries receive the gold dated answer, so the simulation measures how well each trigger picks queries, not how good a real retriever would be. Real performance would also depend on evidence coverage and answer synthesis.

The correctness probe uses the same activation position, linear setup, and fitting procedure as the drift probe, but is trained on correctness instead of drift. It controls for supervised access to hidden states, so the gap between the two probes comes from what each probe reads, not from supervision itself. The hazard cannot rank facts without a holder change before the cutoff (Appendix~\ref{app:hazard}), so on those facts it falls back to random order.
Table~\ref{tab:routing_recall} reports the share of stale facts each trigger catches at a 20\% budget. The drift probe catches 63\%, against 49\% for the hazard and 45\% for first-token entropy.

\begin{table}[h]
\centering
\caption{Stale facts caught at a 20\% verification budget, averaged over models.}
\label{tab:routing_recall}
\small
\begin{tabular}{lc}
\toprule
\bf Trigger & \bf Stale Facts Caught \\
\midrule
Random               & 20\% \\
Semantic entropy     & 24\% \\
Correctness probe    & 29\% \\
First-token entropy  & 45\% \\
Volatility hazard    & 49\% \\
Drift probe          & 63\% \\
\bottomrule
\end{tabular}
\end{table}

Table~\ref{tab:routing_budget} gives end-task accuracy as the budget grows. Entropy needs a 29\% budget to reach the 76.5 that the drift probe has at 20\%, so the probe needs 31\% fewer lookups, or 9 points of absolute budget.
\begin{table}[h]
\centering
\caption{End-task accuracy (\%) by verification budget.}
\label{tab:routing_budget}
\small
\begin{tabular}{lccc}
\toprule
\bf Verification Budget & \bf Random & \bf First-Token Entropy & \bf Drift Probe \\
\midrule
20\% & 71.7 & 73.0 & 76.5 \\
40\% & 78.9 & 80.7 & 83.9 \\
60\% & 86.2 & 88.2 & 90.4 \\
80\% & 93.1 & 94.7 & 95.8 \\
\bottomrule
\end{tabular}
\end{table}

The 63\% already includes facts without a holder change before the cutoff, since the simulation runs on the full test set. Table~\ref{tab:routing_coverage} splits it by history eligibility. On eligible facts, the probe leads the hazard by 12 points. On ineligible facts, the hazard drops to random, and the probe still leads, but its margin over entropy narrows from 19 points to 4. End-task accuracy at the same budget is 77.0 on eligible facts and 71.0 on ineligible ones, which averages to 76.5.

\begin{table}[h]
\centering
\caption{Stale facts caught at a 20\% budget, split by history eligibility.}
\label{tab:routing_coverage}
\small
\begin{tabular}{lccccc}
\toprule
\bf Subset & \bf Share of Queries & \bf Random & \bf Volatility Hazard & \bf First-Token Entropy & \bf Drift Probe \\
\midrule
Eligible    & 92\%  & 20\% & 52\% & 45\% & 64\% \\
Ineligible  & 8\%   & 20\% & 20\% & 44\% & 48\% \\
All queries & 100\% & 20\% & 49\% & 45\% & 63\% \\
\bottomrule
\end{tabular}
\end{table}

\subsection{Adaptive Retrieval Triggers: Protocol and Per-Model Results}
\label{app:rag}

\textbf{Why confidence-based triggers miss drift.}
A drift probe that flags stale answers is only useful if it changes what a system
\emph{does}. The natural use is selective retrieval: query an external source only
when the parametric answer is likely outdated. Our geometry (Section~\ref{sec:orthogonality}) predicts that the triggers we test share a
blind spot: because temporal drift overlaps little with correctness and uncertainty, a
\emph{confidently} stale fact raises no uncertainty flag, so these triggers stay
silent on exactly the queries that most need a refresh.

\textbf{Retrieval as selective prediction.}
We frame retrieval as selective prediction under a fixed budget: each trigger scores
every query, and the top-scoring fraction is sent to retrieval. The measure is recall
of \emph{stale-fact errors}: confidently wrong, post-cutoff answers that a refresh
would fix, measured on the \emph{answerable} slice where the model returns a real
timeline holder, so that errors are drift-induced rather than fabricated. To rule out
leakage, drift scores are computed out-of-fold under fact-disjoint folds, so no
fact is ever scored by a probe trained on it; the uncertainty baselines are
training-free and leakage-free by construction.
\begin{table}[h]
  \centering
  \caption{Stale-fact recall at a $20\%$ retrieval rate.  Answerable slice;
  value $\pm$ half the $95\%$ bootstrap CI; drift scores are out-of-fold,
  fact-disjoint.  Drift-RAG recovers the most stale-fact errors on \emph{every}
  model, beating all adaptive-RAG triggers, including the representation-based methods
  Probing-RAG, CtrlA, and SeaKR.}
  \label{tab:drift-rag}
  \small
  \setlength{\tabcolsep}{4pt}
  \resizebox{0.8\linewidth}{!}{
  \begin{tabular}{llccccr}
  \toprule
  & & \textbf{Qwen-2.5-7B} & \textbf{Llama-3.1-8B} & \textbf{Gemma-2-9B}
    & \textbf{Gemma-2-2B} & \textbf{Mean} \\
  \midrule
  \rowcolor{white}
  \multicolumn{7}{l}{\textit{Output-based}} \\
  \rowcolor{verylightgray}
  & P(True)~\citep{kadavath2022language} & \res{0.38}{0.16} & \res{0.33}{0.07} & \res{0.43}{0.09} & \res{0.13}{0.08} & 0.32 \\
  \rowcolor{white}
  & DRAGIN~\citep{su2024dragin} & \res{0.19}{0.13} & \res{0.35}{0.06} & \res{0.37}{0.08} & \res{0.32}{0.11} & 0.31 \\
  \rowcolor{verylightgray}
  & FLARE~\citep{jiang2023active} & \res{0.16}{0.12} & \res{0.31}{0.06} & \res{0.27}{0.07} & \res{0.26}{0.10} & 0.25 \\
  \rowcolor{white}
  & Semantic entropy~\citep{kuhn2023semantic} & \res{0.16}{0.12} & \res{0.25}{0.06} & \res{0.23}{0.07} & \res{0.25}{0.10} & 0.22 \\
  \midrule
  \rowcolor{white}
  \multicolumn{7}{l}{\textit{Representation-based}} \\
  \rowcolor{verylightgray}
  & Probing-RAG~\citep{baek2025probing} & \res{0.16}{0.12} & \res{0.38}{0.07} & \res{0.31}{0.08} & \res{0.12}{0.08} & 0.24 \\
  \rowcolor{white}
  & SeaKR~\citep{yao-etal-2025-seakr} & \res{0.14}{0.11} & \res{0.27}{0.06} & \res{0.23}{0.07} & \res{0.24}{0.10} & 0.22 \\
  \rowcolor{verylightgray}
  & CtrlA~\citep{huanshuo-etal-2025-ctrla} & \res{0.22}{0.14} & \res{0.19}{0.05} & \res{0.04}{0.04} & \res{0.21}{0.09} & 0.16 \\
  \rowcolor{white}
  & SAPLMA~\citep{azaria2023internal} & \res{0.16}{0.13} & \res{0.46}{0.07} & \res{0.33}{0.08} & \res{0.24}{0.10} & 0.30 \\
  \rowcolor{verylightgray}
  & SEP~\citep{kossen2024semantic} & \res{0.19}{0.13} & \res{0.38}{0.07} & \res{0.43}{0.08} & \res{0.37}{0.11} & 0.34 \\
  \midrule
  \rowcolor{white}
  \multicolumn{7}{l}{\textit{Drift-supervised}} \\
  \rowcolor{white}
  & \textbf{Drift-RAG (ours)} & \best{0.84}{0.15} & \best{0.54}{0.07} & \best{0.65}{0.08} & \best{0.63}{0.11} & \textbf{0.66} \\
  \bottomrule
  \end{tabular}
  }
\end{table}

\textbf{Per-model results.}
The gap over the strongest baseline is largest on Qwen-2.5
($0.84$ vs.\ $0.38$) and smallest on Llama-3.1 ($0.54$ vs.\ $0.46$)
(Table~\ref{tab:drift-rag}). These results suggest that the benefit varies across models but is not captured reliably by the confidence signals evaluated here. The representation-based
competitors fare no better: CtrlA's confidence direction ($0.16$) and SeaKR's
internal eigenscore ($0.22$) sit with the output-confidence methods near the
random-retrieval line, because the directions they extract encode
\emph{confidence} rather than temporal validity, two axes that overlap
little (Section~\ref{sec:orthogonality}). The failure also shows in the errors
themselves: the stale facts Drift-RAG recovers are, in large majority, ones the
entropy trigger leaves behind. In our tests, signals trained on correctness or uncertainty,
supervised or not, do not preferentially surface staleness, and once a drift probe exists, drift-aware retrieval
turns a representational property into a reduction of confidently outdated
answers.

\section{Mechanistic Analysis}
\label{app:mechanism}

\subsection{Activation Patching: Protocol and Per-Layer Profiles}
\label{app:patching}

\paragraph{Protocol.}
Clean and corrupted prompts are byte-identical except for a
year-token substitution (e.g., \emph{In 2016, \ldots} vs.\
\emph{In 2024, \ldots}); the corrupted year selects a different
holder for the same fact.  We locate semantic positions (year,
entity-last, blank, prediction) by character-offset matching and
apply forward pre-hooks at each transformer block to cache clean
residuals, then replace the corrupted residual at one
(layer, position) with the cached clean one.  Year and blank spans
(multi-subtoken under all six tokenizers) are patched jointly.
Recovery is computed per Eq.~(\ref{eq:recovery}).  We require
$|\text{ld}_{\text{clean}} - \text{ld}_{\text{corr}}| \geq 0.5$ to
ensure a measurable retrieval gap.

\paragraph{Three-stage circuit (entity-position models).}
On Llama-2 and Mistral, the recovery trajectory at the entity-last
position reveals: \emph{temporal integration} (L6--L15; attention
heads write year context into the entity representation),
\emph{time-conditioned retrieval} (L16--L19; MLPs select among
stored holders, recovery rises from 0.13 to 1.29 on Llama-2), and
a \emph{post-retrieval plateau} (L19--L28; recovery exceeds 1.0,
indicating the patched representation amplifies the clean answer's
logit beyond the original clean run).  Late answer-extraction heads
(L29--L31) transfer the selected representation to the prediction
position.  On prediction-position models, recovery rises
monotonically (Llama-3.1: 0.04 at L13, 0.97 at L31).

\paragraph{Year-token decay.}
Recovery at the year token follows the expected causal pattern. It is 1.000 at L0, because replacing the embedding makes the corrupted prompt identical to the clean one, stays at or above 1.0 through L15, and then falls below 0.05 from L19 onward, as temporal information moves to the entity and last-token positions.

\subsection{DLA Correlations Excluding Layer 0}
\label{app:dla_no_l0}

The Gemma L0 peak in Table~\ref{tab:dla} reflects a known DLA
artifact under tied
embeddings~\citep{janiak2024adversarial}. Table~\ref{tab:dla_no_l0} recomputes the \textsc{Stale-Recall} vs.\ \textsc{Confabulation} trajectory correlations with L0 excluded.

\begin{table}[h]
\centering
\caption{Correlation between the per-layer DLA curves of \textsc{Stale-Recall} and \textsc{Confabulation}, with all layers and with layer 0 excluded.}
\label{tab:dla_no_l0}
\scriptsize
\begin{tabular}{lcc}
\toprule
\textbf{Model} & \bf $r$ (all layers) & \bf $r$ ($\ell \geq 1$) \\
\midrule
Llama-2    & 0.986 & 0.984 \\
Mistral    & 0.947 & 0.943 \\
Llama-3.1  & 0.883 & 0.880 \\
Qwen-2.5   & 0.815 & 0.811 \\
Gemma-2-9B & 0.993 & 0.978 \\
Gemma-2-2B & 0.997 & 0.968 \\
\bottomrule
\end{tabular}
\end{table}

All correlations remain above 0.81, so the similarity of the curves does not come from layer 0.

\subsection{Causal Steering: Full Tables and Controls}
\label{app:steering_full}

\paragraph{Suppression across states.}
Table~\ref{tab:steering_suppression} reports how many outputs change when we remove $\hat{\mathbf{d}}$ from the hidden states.

\begin{table}[h]
\centering
\caption{Outputs that change when we remove the steering direction $\hat{\mathbf{d}}$, by state. Each entry gives the changed outputs out of the tested queries.}
\label{tab:steering_suppression}
\scriptsize
\rowcolors{2}{verylightgray}{white}
\begin{tabular}{lcccc}
\toprule
\textbf{Model} & \shortstack{\textbf{Stale-}\\\textbf{Recall}} & \shortstack{\textbf{Stable-}\\\textbf{Correct}} & \textbf{Confabulation} & \textbf{Total} \\
\midrule
Llama-2    & 0/100 (0\%)  & 0/100 (0\%)  & 0/100 (0\%)   & 0/300 \\
Gemma-2-2B & 0/37 (0\%)   & 4/100 (4\%)  & 6/100 (6\%)   & 10/237 \\
Mistral    & 5/58 (9\%)   & 2/100 (2\%)  & 16/100 (16\%) & 23/258 \\
\rowcolor{white}
Llama-3.1  & 7/75 (9\%)   & 2/100 (2\%)  & 12/100 (12\%) & 21/275 \\
\bottomrule
\end{tabular}
\end{table}

Removing $\hat{\mathbf{d}}$ changes no outputs on Llama-2 and at most 9\% of all outputs on the other models, with the most changes on \textsc{Confabulation} (up to 16\%).
\paragraph{Amplification on \textsc{Stable-Correct} and
\textsc{Confabulation}.}
On \textsc{Stable-Correct}, $+\alpha\hat{\mathbf{d}}$
amplification decreases the output holder's logit while increasing
past holders' logits on three of five models.  On
\textsc{Confabulation}, the fabricated answer's logit decreases
while the current holder's logit increases on four of five models.
Both patterns confirm that $\hat{\mathbf{d}}$ encodes holder
competition rather than staleness specifically.

\paragraph{Reverse steering.}
$-\alpha\hat{\mathbf{d}}$ produces a qualitatively different
pattern: both stale and current holders' logits decrease, with
probability mass redistributing toward the broader vocabulary,
confirming structured causal connectivity rather than random
perturbation.

\paragraph{DoM confound mitigation.}
We compute $\hat{\mathbf{d}}$ within the
\texttt{head\_of\_government} relation and over the
\textsc{Stale-Recall} vs.\ \textsc{Stable-Correct} contrast only.
The \S\ref{sec:dom_dissociation} concern that DoM directions are
confounded with correctness applies to DoM computed across all
cells; restricting to a single relation and a single binary contrast
reduces the cosine of $\hat{\mathbf{d}}$ with $\mathbf{w}_c$ and
$\mathbf{w}_u$ to ${\leq}\,0.18$ on every model (vs.\
${\leq}\,0.92$ for unrestricted DoM).

\textbf{Probe and Steering Directions.}
The drift probe $\mathbf{w}_d$ and the steering direction $\hat{\mathbf{d}}$ are different objects: $\mathbf{w}_d$ is the supervised probe we use for detection, and $\hat{\mathbf{d}}$ is the difference-of-means direction we amplify in Section~\ref{sec:steering}. Table~\ref{tab:probe_steer_cos} measures both at the same token position and layer used for the interventions. The right column pools the contrast across all relations instead of restricting it to \texttt{head\_of\_government}. The overlap is small either way, under 2\% of shared variance on every model, and the mean is 0.042 at the cross-validated layers of Section~\ref{sec:probes}, so it does not depend on the layer protocol. We therefore report two separate findings: $\mathbf{w}_d$ reads staleness, and $\hat{\mathbf{d}}$ shifts probability between competing holders.

\begin{table}[h]
\centering
\caption{Absolute cosine between the drift probe $\mathbf{w}_d$ and the steering direction $\hat{\mathbf{d}}$ at the steering layer of each model. The steering direction is computed within \texttt{head\_of\_government} (as in Section~\ref{sec:steering}) or pooled over all relations.}
\label{tab:probe_steer_cos}
\small
\begin{tabular}{lccc}
\toprule
 & & \multicolumn{2}{c}{\bf Absolute Cosine with the Drift Probe} \\
\cmidrule(lr){3-4}
\bf Model & \bf Steering Layer & \bf One Relation & \bf All Relations \\
\midrule
Llama-2   & L24 & 0.103 & 0.135 \\
Mistral   & L6  & 0.055 & 0.039 \\
Llama-3.1 & L29 & 0.045 & 0.078 \\
Qwen-2.5  & L10 & 0.069 & 0.044 \\
Gemma-9B  & L33 & 0.060 & 0.054 \\
Gemma-2B  & L13 & 0.048 & 0.029 \\
\midrule
Mean      &     & 0.063 & 0.063 \\
\bottomrule
\end{tabular}
\end{table}

\clearpage
\section*{NeurIPS Paper Checklist}

\begin{enumerate}

\item {\bf Claims}
    \item[] Question: Do the main claims made in the abstract and introduction accurately reflect the paper's contributions and scope?
    \item[] Answer: \answerYes{}
    \item[] Justification: The abstract and introduction state our contributions (dataset, linear drift probe and its independence from correctness and uncertainty, selective retrieval), and the scope is accurately bounded to six instruction-tuned models, four relations, and $2$--$9$B parameter range. Appendix~\ref{app:relations} extends the scope to 18 relations.

\item {\bf Limitations}
    \item[] Question: Does the paper discuss the limitations of the work performed by the authors?
    \item[] Answer: \answerYes{}
    \item[] Justification: Limitations are discussed in ~\ref{sec:conclusion} covering: four Wikidata relations only, $2$--$9$B parameter range, unknown scaling behavior, and the requirement for drift-labeled examples to train the probe. Section~\ref{sec:hazard} also reports where the probe is weakest, on facts with little temporal evidence in pretraining.

\item {\bf Theory assumptions and proofs}
    \item[] Question: For each theoretical result, does the paper provide the full set of assumptions and a complete (and correct) proof?
    \item[] Answer: \answerNA{}
    \item[] Justification: The paper makes no formal theoretical claims; all results are empirical, supported by five complementary independence measures and bootstrap confidence intervals.

\item {\bf Experimental result reproducibility}
    \item[] Question: Does the paper fully disclose all the information needed to reproduce the main experimental results of the paper to the extent that it affects the main claims and/or conclusions of the paper (regardless of whether the code and data are provided or not)?
    \item[] Answer: \answerYes{}
    \item[] Justification: Model checkpoints (Table~\ref{tab:models}), decoding settings (greedy decoding, at most 25 new tokens, 16-bit precision), probe architecture (L1-regularized ISTA linear probe), train/test split by fact, and the full state-assignment procedure (73 unit tests, Appendix~\ref{app:cell_assignment}) are all fully specified. We choose the layer and $\lambda$ jointly by 3-fold fact-grouped cross-validation on training facts (Section~\ref{sec:probes}).

\item {\bf Open access to data and code}
    \item[] Question: Does the paper provide open access to the data and code, with sufficient instructions to faithfully reproduce the main experimental results, as described in supplemental material?
    \item[] Answer: \answerYes{}
    \item[] Justification: The dataset and code, including the state-assignment pipeline validated by 73 unit tests and the query procedure against our Wikidata snapshot (April 27, 2026), are released at \url{https://github.com/mbzuai-nlp/geometry-of-staleness}.

\item {\bf Experimental setting/details}
    \item[] Question: Does the paper specify all the training and test details (e.g., data splits, hyperparameters, how they were chosen, type of optimizer) necessary to understand the results?
    \item[] Answer: \answerYes{}
    \item[] Justification: Section~\ref{sec:probes} specifies all probing details: activation extraction position, L1/L2 regularization, $\lambda$ grid, class-balanced oversampling, layer sweep, and the use of only query years after each model's cutoff. We choose the layer by cross-validation on training facts.

\item {\bf Experiment statistical significance}
    \item[] Question: Does the paper report error bars suitably and correctly defined or other appropriate information about the statistical significance of the experiments?
    \item[] Answer: \answerYes{}
    \item[] Justification: All AUROCs are reported with 500-resample stratified bootstrap 95\% CIs; cross-model comparisons use one-sided Mann--Whitney $U$ with $\alpha=0.05$; score correlations use Pearson $r$ with 1,000-resample bootstrap CIs.

\item {\bf Experiments compute resources}
    \item[] Question: For each experiment, does the paper provide sufficient information on the computer resources (type of compute workers, memory, time of execution) needed to reproduce the experiments?
    \item[] Answer: \answerYes{}
    \item[] Justification: Table~\ref{tab:models} states all models are loaded in float16 on a single A100; the probing setup and activation extraction procedure are fully described, making compute requirements straightforward to estimate.

\item {\bf Code of ethics}
    \item[] Question: Does the research conducted in the paper conform, in every respect, with the NeurIPS Code of Ethics \url{https://neurips.cc/public/EthicsGuidelines}?
    \item[] Answer: \answerYes{}
    \item[] Justification: The paper studies temporal knowledge drift in publicly available instruction-tuned models using Wikidata, a public knowledge base. No human subjects, sensitive data, or dual-use risks are involved.

\item {\bf Broader impacts}
    \item[] Question: Does the paper discuss both potential positive societal impacts and negative societal impacts of the work performed?
    \item[] Answer: \answerYes{}
    \item[] Justification: The work enables detection of outdated LLM outputs in high-stakes deployments (positive); no significant negative societal impact is identified, as the probe requires drift-labeled data and cannot be trivially misused. ~\ref{sec:conclusion} discusses one risk: a system that relies on the probe can miss outdated answers where its signal is weak.

\item {\bf Safeguards}
    \item[] Question: Does the paper describe safeguards that have been put in place for responsible release of data or models that have a high risk for misuse (e.g., pre-trained language models, image generators, or scraped datasets)?
    \item[] Answer: \answerNA{}
    \item[] Justification: The released assets are a Wikidata-derived query dataset and linear probe weights; neither poses meaningful misuse risk.

\item {\bf Licenses for existing assets}
    \item[] Question: Are the creators or original owners of assets (e.g., code, data, models), used in the paper, properly credited and are the license and terms of use explicitly mentioned and properly respected?
    \item[] Answer: \answerYes{}
    \item[] Justification: All six model checkpoints are cited with their HuggingFace identifiers (Table~\ref{tab:models}); Wikidata is used under its CC0 license; all baseline methods are cited. We use the models under their licenses: the Llama~2 and Llama~3.1 Community Licenses, Apache~2.0 for Mistral and Qwen-2.5, and the Gemma Terms of Use. We use Dolma under ODC-BY.

\item {\bf New assets}
    \item[] Question: Are new assets introduced in the paper well documented and is the documentation provided alongside the assets?
    \item[] Answer: \answerYes{}
    \item[] Justification: The dataset of 3,511 queries over 1,342 facts is fully documented: construction procedure (Section~\ref{sec:drift_def}), query pipeline, state-assignment logic (Appendix~\ref{app:cell_assignment}), and per-model state counts (Table~\ref{tab:cell_dist}) are all provided. Appendix~\ref{app:relations} documents the 18-relation extension.

\item {\bf Crowdsourcing and research with human subjects}
    \item[] Question: For crowdsourcing experiments and research with human subjects, does the paper include the full text of instructions given to participants and screenshots, if applicable, as well as details about compensation (if any)?
    \item[] Answer: \answerNA{}
    \item[] Justification: The paper involves no crowdsourcing or human subjects; all data is sourced from Wikidata and model outputs are generated automatically.

\item {\bf Institutional review board (IRB) approvals or equivalent for research with human subjects}
    \item[] Question: Does the paper describe potential risks incurred by study participants, whether such risks were disclosed to the subjects, and whether Institutional Review Board (IRB) approvals (or an equivalent approval/review based on the requirements of your country or institution) were obtained?
    \item[] Answer: \answerNA{}
    \item[] Justification: No human subjects are involved.

\item {\bf Declaration of LLM usage}
    \item[] Question: Does the paper describe the usage of LLMs if it is an important, original, or non-standard component of the core methods in this research? Note that if the LLM is used only for writing, editing, or formatting purposes and does \emph{not} impact the core methodology, scientific rigor, or originality of the research, declaration is not required.
    \item[] Answer: \answerNA{}
    \item[] Justification: LLMs are the object of study, not a component of our method. The six evaluated models are fully identified in Table~\ref{tab:models}. Any LLM use for writing or formatting does not impact the methodology.

\end{enumerate}

\end{document}